\documentclass[a4paper,fleqn]{cas-dc}

\usepackage[numbers]{natbib}
\usepackage{amssymb}
\usepackage{amsmath,amsfonts}
\usepackage{algorithmic}
\usepackage{algorithm}
\usepackage{array}
\usepackage[caption=false,font=footnotesize,labelfont=sf,textfont=sf]{subfig}
\usepackage{textcomp}
\usepackage{stfloats}
\usepackage{url}
\usepackage{verbatim}
\usepackage{graphicx}
\def\BibTeX{{\rm B\kern-.05em{\sc i\kern-.025em b}\kern-.08em
    T\kern-.1667em\lower.7ex\hbox{E}\kern-.125emX}}
\usepackage{color}
\usepackage{balance}
\usepackage{pifont}
\usepackage{tabularx}
\usepackage{booktabs}
\usepackage{multirow}
\usepackage{etoolbox}
\usepackage{bbm}
\usepackage{bm}
\makeatletter
\patchcmd{\@makecaption}
  {\scshape}
  {}
  {}
  {}
\makeatletter
\patchcmd{\@makecaption}
  {\\}
  {.\ }
  {}
  {}

\makeatother

\def\tsc#1{\csdef{#1}{\textsc{\lowercase{#1}}\xspace}}
\tsc{WGM}
\tsc{QE}

\begin{document}
\let\WriteBookmarks\relax
\def\floatpagepagefraction{1}
\def\textpagefraction{.001}
\let\printorcid\relax 

\shorttitle{TSA-MLT}    

\shortauthors{Fei Guo et al.}

\title[mode = title]{Task-Specific Alignment and Multiple-level Transformer for Few-Shot Action Recognition}

\author[1]{Fei Guo}[type=editor,
    auid=000]
\fnmark[1] 
\ead{co.fly@stu.xjut.edu.cn}

\author[1]{Li Zhu}
\ead{zhuli@xjtu.edu.cn} 
\fnmark[2] 
\cormark[1] 

\author[1]{YiKang Wang}
\ead{funnyQ@stu.xjtu.edu.cn}
\fnmark[2] 

\author[2]{Jing Sun}
\ead{jing.sun1@siat.ac.cn}
\fnmark[3]

\address[1]{School of Software Engineering, Xi'an Jiaotong University, China}
\address[2]{Shenzhen Institutes of Advanced Technology, Chinese Academy of Sciences, China}

\cortext[1]{Corresponding author}


\maketitle
\begin{abstract}
In the research field of few-shot learning, the main difference between image-based and video-based is the additional temporal dimension. In recent years, some works have used the Transformer to deal with frames, then get the attention feature and the enhanced prototype, and the results are competitive. 
However, some video frames may relate little to the action, and only using single frame-level or segment-level features may not mine enough information.
We address these problems sequentially through an end-to-end method named “Task-Specific Alignment and Multiple-level Transformer Network (TSA-MLT)". 
The first module (TSA) aims at filtering the action-irrelevant frames for action duration alignment. Affine Transformation for frame sequence in the time dimension is used for linear sampling.
The second module (MLT) focuses on the Multiple-level feature of the support prototype and query sample to mine more information for the alignment, which operations on different level features.
We adopt a fusion loss according to a fusion distance that fuses the L2 sequence distance, which focuses on temporal order alignment, and the Optimal Transport distance, which focuses on measuring the gap between the appearance and semantics of the videos. 
Extensive experiments show our method achieves state-of-the-art results on the HMDB51 and UCF101 datasets and a competitive result on the benchmark of Kinetics and something-2-something V2 datasets. Our code will be available at the URL: \url{https://github.com/cofly2014/tsa-mlt.git}.
\end{abstract}



\section{Introduction} \label{introduction}
Video action recognition is an important research area. The rapid development of video action recognition based on deep learning is due to the use of a large amount of labeled videos\cite{simonyan2014two,tran2015learning,wang2016temporal-tsn,tcsvt-fs4,tcsvt-fs5}.
However, the collection of video labels is time-consuming and labor-expensive. Consequently, the few-shot action recognition has received more and more attention due to the significantly reduced demand for labeled samples. 
For few-shot action recognition, earlier works tried to aggregate the representation of the support video in each class and get the prototype. 
As the distribution of actions is not the same in different videos, the action speed is different, the critical content occurs at different time points, and the action duration is often misaligned, it can be seen that some aggregation methods ignore the important temporal information of videos. 
Later, researchers introduced the attention mechanism and Transformer to address the spatial and temporal alignment, such as the TARN\cite{bishay2019tarn},  ARN\cite{zhang2020few-ARN}, TAEN\cite{ben2021taen}, TRX\cite{perrett2021temporal-trx}, STRM\cite{thatipelli2022spatio-strm}.
Among these works, in 2021, TRX\cite{perrett2021temporal-trx} selects a part-combination of tuples and gets the attention of specific tuples. Sometimes, there are several Transformer instances in TRX; for example, one Transformer branch extracts features from tuples that contain two frames, and another Transformer branch extracts features from tuples that contain three frames, etc. It averages the class probability of the query feature from each Transformer branch and gets the final class probability. \textbf{Our work gets inspiration from TRX \cite{perrett2021temporal-trx} and STRM \cite{thatipelli2022spatio-strm}, and adopts them as the baseline.}
Like most few-shot action recognition models, for the TRX, the first step is frame sampling. Since the probability distribution of video action is various and the sampling operation is random, the action location duration in the sampled frames is not the same. We need to align the action location duration.
Also, this work only focuses on the alignment of frame-to-frame or segment-to-segment (The segments for comparison contain the same number of frames), which are single-level features. However, the relationship between representations of different levels is perhaps also significant. 
We can see that aligning the content for the video is still challenging, as the action location alignment and Multiple-level feature alignment remain under-explored.

\begin{figure} 
\centering
\includegraphics[width=8cm]{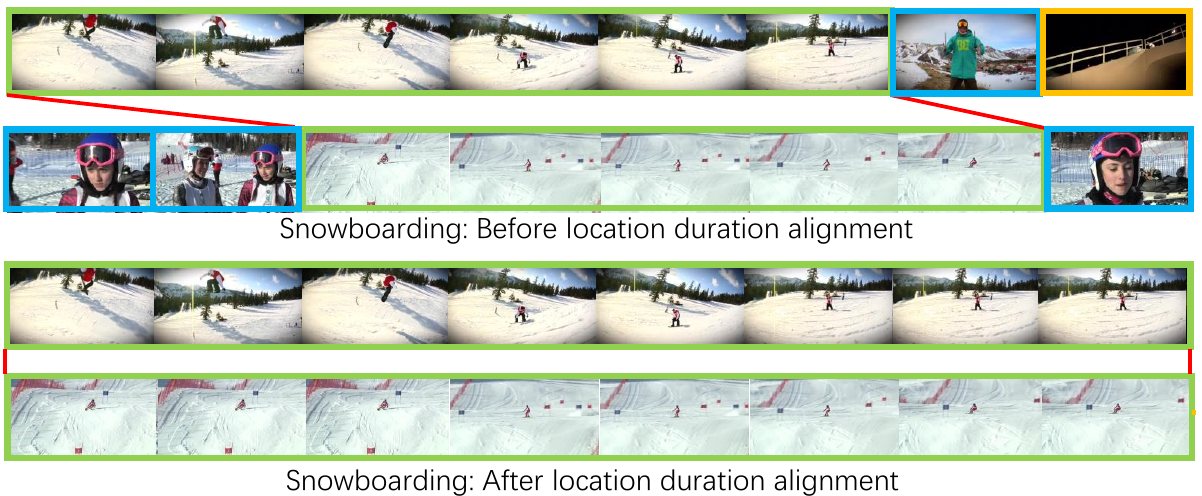}
\caption{\label{fig:alignment}The illustration shows the location duration alignment. At the beginning and end, sometimes, a few frames are less important or have misleading information. In the illustration, the frames with a blue border have less important information related to snowboarding, and the frame with a yellow border perhaps has misleading information related to snowboarding. Using the TSA, we aim to filter this information forcibly. For the action ``snowboarding", the upper is before the alignment, and the lower is after the alignment.}
\label{fig:1}
\end{figure}

\begin{figure*}[htbp]
    \centering
    \subfloat[Two sequences of video frames that need to be aligned ]{
    \label{fig:2.a}
    \includegraphics[width=0.8\textwidth]{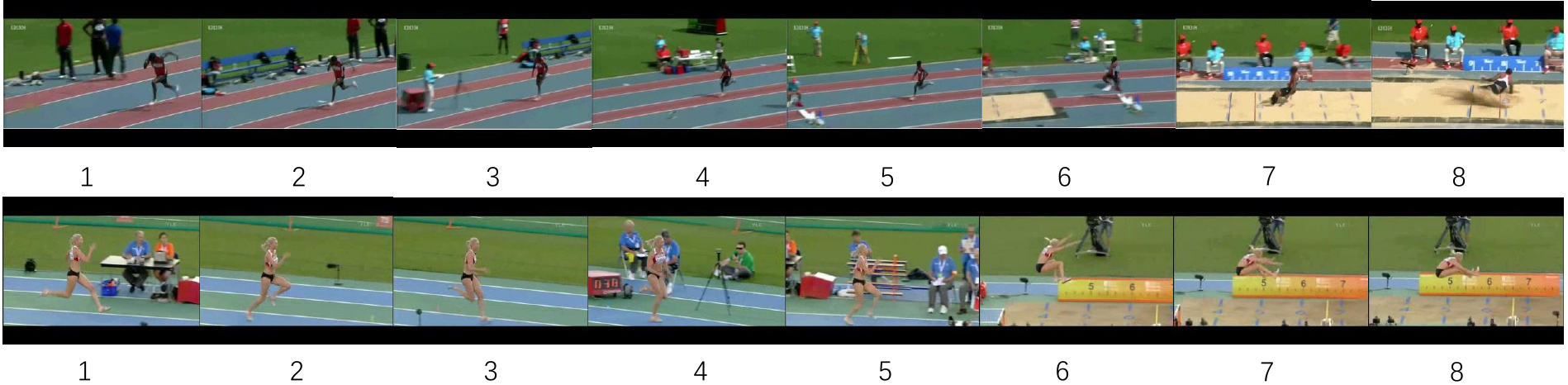}
    }  
   
    \subfloat[Examples of Representations alignment]{
    \label{fig:2.b}
    \includegraphics[width=0.8\textwidth]{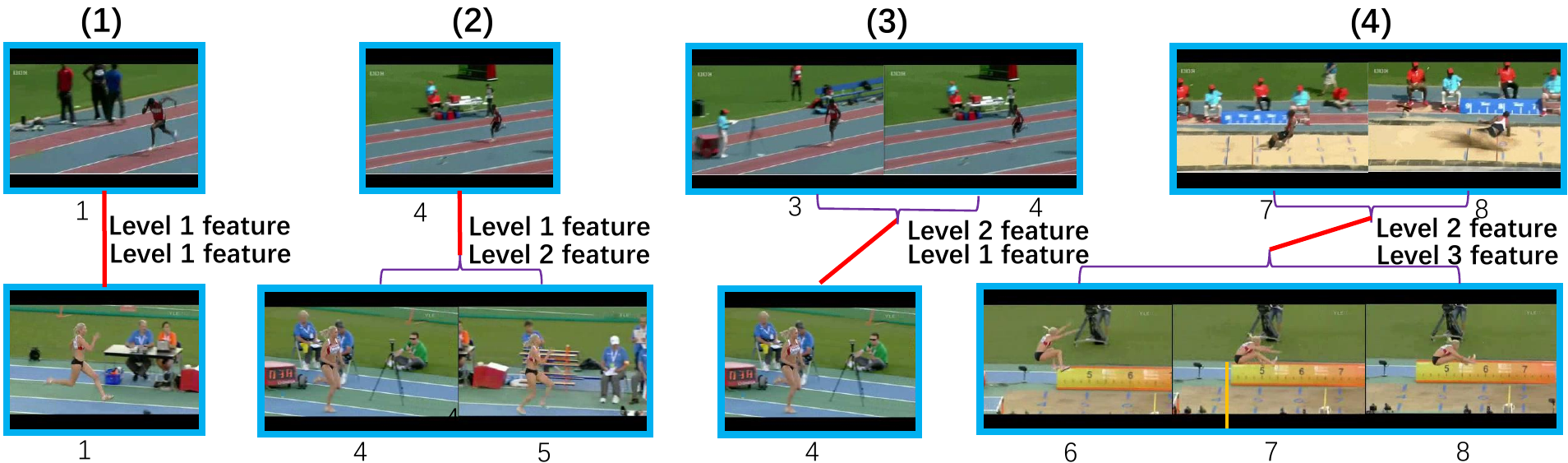}
    }
    \caption{Mutiple-level alignment.
 (a) It shows two sequences of video frames that need to be aligned. (b) (1) A single frame in the upper video to a single frame in the lower video. (2) A single frame in the upper video to the representation of level 2 in the lower video. (3) The representation of level 2 in the upper video to a single frame in the lower video. (4) The representation of level 2 in the upper video to the representation of level 3 in the lower video.
 }
 \label{fig:2}
\end{figure*}

To cope with the above issues, we devise a \textbf{``Task-Specific Alignment and Multiple-level Transformer Network (TSA-MLT)"} for few-shot action recognition. 
The first module is \textbf{Task-Specific Alignment (TSA)}. Action location duration of videos often has a misalignment issue, and sometimes a few frames are less critical or misleading (see Figure \ref{fig:1}). 
With Zoom and Pan Variables generated from a Position network based on CNN and an Affine Transformation in time dimension under such variables, our TSA can adjust the action location duration from two directions. The filtering strategy is fixed when the Position network is well-trained. Still, in few-shot action recognition, tasks usually cause large inter-task variances, so using a fixed network to deal with the problem seems insufficient to always conduct good features. To this end, we adopt a Task-Specific learner, which generates parameters for adjusting the zoom and pan. 
The second module is \textbf{Multiple-level Transformer (MLT)}. This module aims for alignment between the Multiple-level features. The Multiple-level features contain tuples composed of different numbers of frames. Using the Transformer mechanism, a query-specific class prototype of Multiple-level is constructed to match each Multiple-level sequence of the query set. The feature alignment from different levels (containing different numbers of frames) can make the model more robust. Figure \ref{fig:2} shows some examples of alignment for the multiple-level features. Unlike the TRX-related works that need to create several Transformer instances based on the selection of cardinalities for the tuples, our MLT only requires one Transformer instance. 
We use a network for each cardinality to aggregate all the origin tuples into a small number of tuple representations. This way, we avoid the feature dimension being too large and the model overfitting. 
In addition to the two modules mentioned above, we use both the sequence distance for mapping the tuples sequences and the Optimal Transport distance to focus on the appearance and semantics. Furthermore, we design a simple yet effective fusion network to combine the two distances.

Our contribution can be summarized as follows:
\begin{itemize}
\item{(1) We propose a Task-Specific Alignment (TSA) module for filtering less important or misleading information in order to obtain an appropriate action location duration alignment.
}
\item{(2) We design a Multiple-level Transformer (MLT) module that uses Multiple-level features from different cardinalities to create query-specific prototypes for Multiple-level feature alignment.
}
\item{(3) In our work, the Optimal Transport theory is introduced to implement the appearance and semantic matching for the Multiple-level features.
}
\item{(4) We contact the distance of $l2$ between two sequences and the distance of Optimal Transport element-wise, then use a network for a fusion distance, which is used for the cross-entropy loss. 
}
\item{(5) Extensive experiments have been conducted, and the results on four benchmarks, HMDB51, UCF101, Kinetics, and Something-Something-V2, demonstrate that our TSA-MLT Network has a favorable performance and can compete with the state-of-the-art few-shot action recognition methods.
}
\end{itemize}

\section{Related Work}
\subsection{Few-shot learning}
The few-shot learning refers to using data samples that are far fewer than required for deep learning to achieve close to or even exceed the effect of deep learning, which uses labeled large samples. In recent years, as the investigation continues, researchers classify few-shot learning into three categories: (1) Metric-based methods \cite{simon2020adaptive-few_shot_learning1,snell2017prototypical_few_shot_learning2,sung2018learning-relation-networkfew_shot_learning3,melekhov2016-siamese—few_shot_learning4,vinyals2016matching—few_shot_learning5,liu2018learning-TNP-few_shot_learning7, ye2020few-embedding-adaptation-few_shot_learning8,tcsvt-fs1,tcsvt-fs2,kbs-zhou2022dynamic-metric}
(2) Model-based methods \cite{finn2017model-MAML-few-shot-model1,rusu2018-metaL-few-shot-model3,ravi2017optimizationL-few-shot-model4,9999670-generalization-of-MAML,kbs-feng2022meta, kbs-zheng2023detach-meta-transfer} 
(3) Data-Augmentation based methods\cite{ratner2017learning-few-augmentation2,chen1804semantic-few-augmentation3,perez2017effectiveness-few-augmentation1,pahde2021multimodal-few-augmentation4,10035001-generation-few-shot-learning,tcsvt-fs3}. For example, Siamese Network\cite{melekhov2016-siamese—few_shot_learning4} measures the similarity for a pair of images, Prototypical Network \cite{snell2017prototypical_few_shot_learning2} computes the distances of Euclidean between the prototypes and queries, and MAML\cite{finn2017model-MAML-few-shot-model1} is a meta-learning strategy that intends to learn a good model initialization of the parameters for a network that can rapidly adapt to new tasks. Multimodal few-shot task\cite{pahde2021multimodal-few-augmentation4} can generate extra images based on the text description to compensate for the lack of data.

\subsection{Few-shot action recognition}
Since 2018, many methods of few-shot action recognition have emerged. We can divide these works into the following categories: (1) Memory-based methods. CMN\cite{zhu2018compound}, and CMN-J\cite{zhu2020label} belong to this category, which encodes the video representation using multi-significant embedding, and the key and values are stored in a memory and can be updated easily. ``Few-shot action recognition with cross-modal memory network"\cite{zhang2020few1} is also based on memory, and this work introduces the cross-modal: one modal is the video content, and another modal is the action label. SMFN\cite{qi2020few-smfn} proposes to solve the few-shot video classification problem by learning a set of SlowFast networks enhanced by memory units. (2) Data-argument-based methods. ProtoGAN\cite{kumar2019protogan} uses a generative adversarial network with specific semantics as the condition to synthesize additional samples of novel classes for few-shot learning. AMeFu-Net\cite{fu2020depth} is also about data augmentation using depth features to enhance the information. (3) Metric-based methods for order matching. OTAM \cite{cao2020few-otam} computes the distance matrix following the DTW\cite{muller2007dynamic} method and makes a strict alignment. TARN\cite{bishay2019tarn} first introduced the attention mechanism for temporal alignment, and its alignment is frame level. TRX\cite{perrett2021temporal-trx} uses the Cross-Transformer to deal with frame tuples, then gets the tuples' feature embedding alignment. The matching method of STRM \cite{thatipelli2022spatio-strm} is similar to the TRX and adds some pre-processing for feature enrichment. ARN \cite{zhang2020few-ARN} focuses on robust similarity, spatial and temporal modeling of short-term and long-term range action patterns via permutation-invariant pooling and attention. FAN\cite{tan2019learning-FAN} encodes videos and incorporates image features for comparison with an element-wise combination. (4) Metric-based methods for appearance and semantics. PAL\cite{zhu2021few-PAL} proposes a prototype-center loss and hybrid attention for order matching. CMOT\cite{lu2021few-cmot} uses the Optimal Transport theory to compare the content of the video, not only focusing on the ordered alignment but also focusing on the video semantics. The HCR\cite{li2022hierarchical-HCR} uses a Wasserstein distance to match the sequence of two sub-actions. ``Few-Shot Learning of Video Action Recognition Only Based on Video content"\cite{bo2020few-onlybasedvideocontent} extracts the global video-level representation through aggregation.  

Our work belongs to the category of metric-based methods for order matching and involves metric-based methods for appearance and semantics.

\section{Methodology}
\subsection{Problem Setting}
In the field of few-shot action recognition, the video datasets are split into  $D_{train}$, $D_ {test}$, $D_ {val}$, all the split datasets should be disjoint, which means there are no overlapping classes between each split dataset. In the $D_{train}$, it contains abundant labeled data for each action class, while there are only a few labeled samples in the $D_ {test}$, and the $D_ {val}$ is used for model evaluation during the training episode. No matter $D_{train}$, $D_ {test}$, or $D_ {val}$, they all follow a standard episode rule. Many episodes are also called tasks during the training, testing, or validation. 
In each episode, $N$ classes with each class containing $K$ samples in $D_{train}$, $D_ {test}$, or $D_ {val}$ are sampled as ``support set". The samples from the rest videos under certain classes of each split DB are sampled as ``query set," just as $P$ samples are selected from N classes to construct the ``query set". The goal of few-shot action recognition is training a model using $D_{train}$, which can be generalized well to the novel classes in the $D_ {test}$ only using $N\times K$ samples in the support set of $D_ {test}$. Let $q = ({q_1, q_2, \cdots , q_m })$ represents a query video with $m$ uniformly sampled frames. We use $C$ to represent the set of classes, $C =\{c_1,\cdots, c_N\}$, and the goal is to classify a query video $q$ into one of the classes $c_i \in C$. In our work, the support set is defined as $S$, and the query set is defined as $Q$. For the class $c$, the support set $S_c$  can be expressed as $S_c = \{S_c^{1}, \cdots ,S_c^{k} \cdots, S_{c}^{K}\}$, where $S_{c}^{k}= ({S_c^{k1}, \cdots, S_{c}^{km}})$, $1 \leq k \leq K$, $m$ is the number of frames.

\subsection{Overall framework}

\begin{figure*}
\centering
\includegraphics[width=18cm]{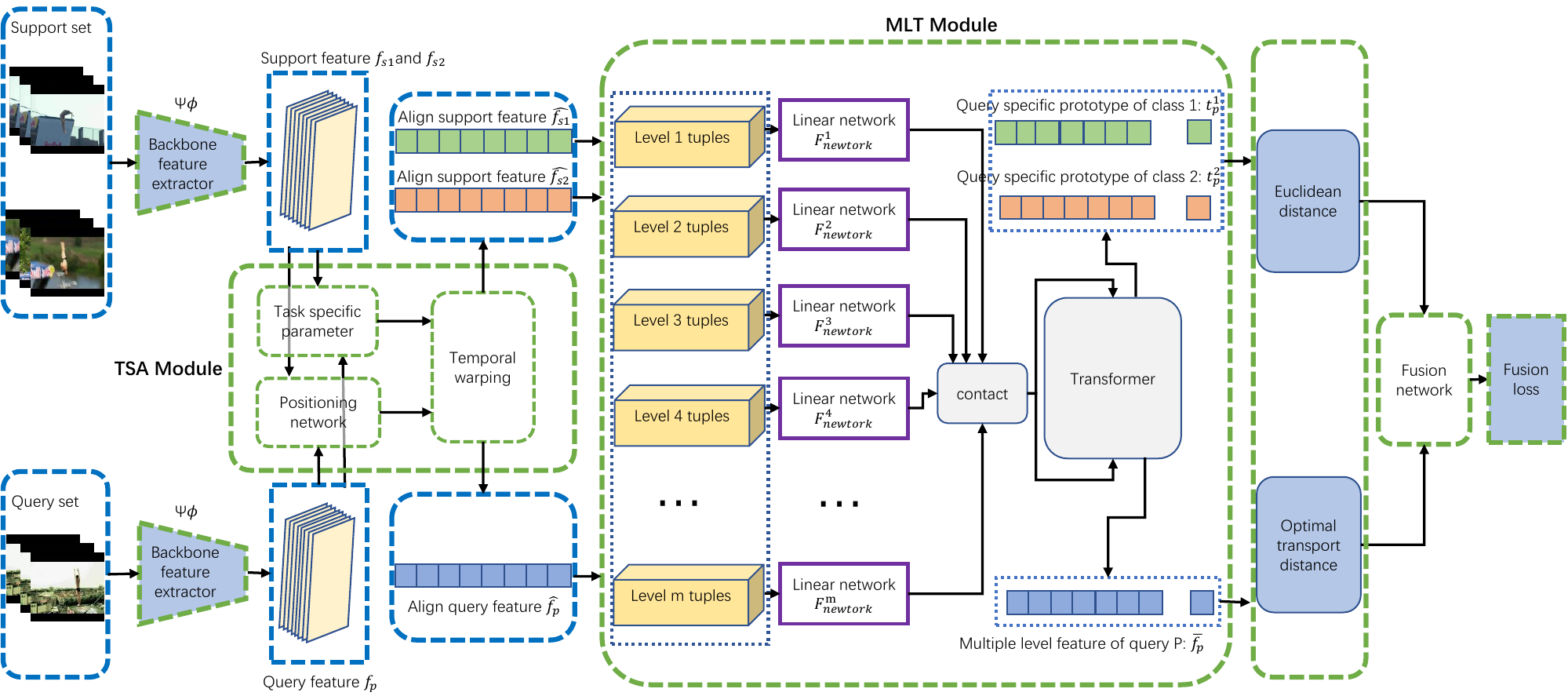}
\caption{\label{fig:framework_of_paper}The Framework of TSA-MLT. In this illustration, the visual workflow of our model can be seen as: \textbf{(a)} Backbone of RestNet50 ($\Psi_{\phi}$).
\textbf{(b)} Task-Specific Alignment(TSA). 
\textbf{(c)} Multiple-level Transformation(MLT). 
\textbf{(d)} Sequence order distance and OT distance.
\textbf{(e)} Simple fusion network for the two distances. 
(Note: The green dashed areas represent the modules, while the blue dashed areas represent features in each step. For simplicity, we use a 2-way 1-shot setting in the illustration.)
} 
\label{fig:3}
\end{figure*}

We propose a framework for few-shot action recognition as shown in Figure \ref{fig:3}. Our work consists of five parts: 
(a)  an Embedding Network $\Psi_{\phi}$ which extracts embedding features using the pre-training model on an enormous dataset; 
(b) a Task-Specific Alignment module (TSA), which is used for filtering out the less essential or misleading information to get a location duration alignment; 
(c) a Multiple-level Transformer module (MLT), with only one Transformer instance, it can aggregate different level features, then the information of different levels is used to create the enhanced prototype, which is used for Multiple-level feature alignment; 
(d) two types of distance: Sequence mapping distance, which is order sensitive, and Optimal Transport distance, which is focused on minimizing the semantic gap between support and query; 
(e) a network for fusing the Sequence distance and Optimal Transport distance, the output of the network can obtain complementary information between two distances and will be used for computing the cross-entropy loss.

\subsection{Task-Specific Alignment module(TSA)}
\begin{figure}
\centering
\includegraphics[width=8.6cm]{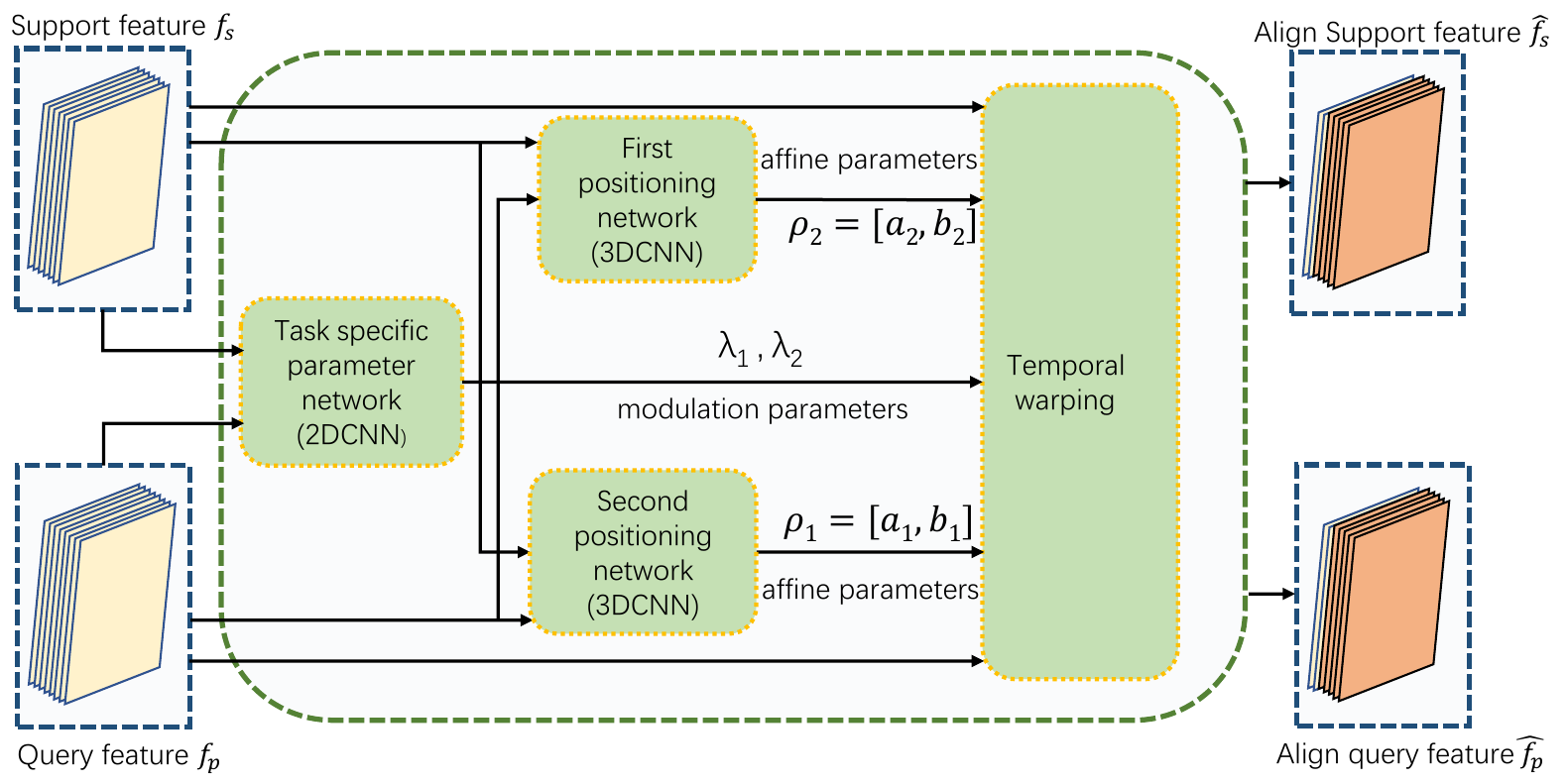}
\caption{\label{fig:TSA}The Task-Specific Alignment. Features are sent into the first and second positioning network, which generates the parameters for warping. The Task-Specific parameter network generates the modulation parameters that assign the degree of two warping transformations.}
\label{fig:4}
\end{figure}

\begin{figure*}[htbp]
    \centering
    \subfloat[Task-Specific parameter network]{
    \label{fig:4_plus.a}
    \includegraphics[width=0.5\textwidth]{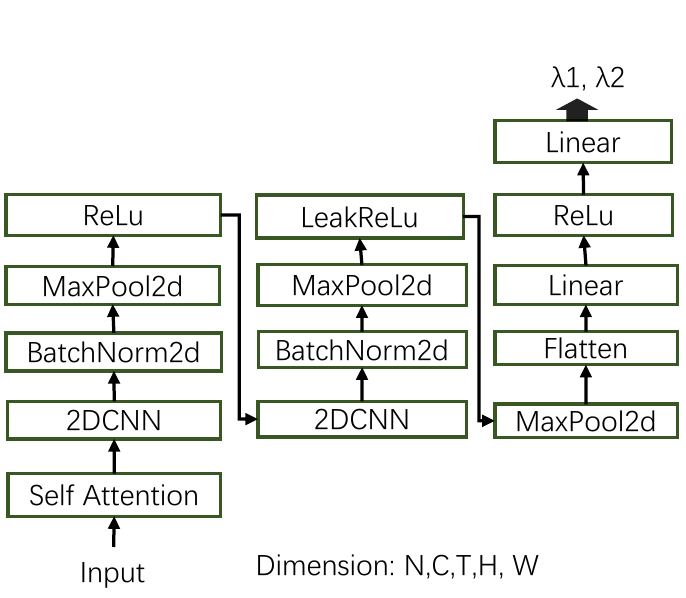}
    }
     \centering
    \subfloat[Positioning network]{
    \label{fig:4_plus.b}
    \includegraphics[width=0.5\textwidth]{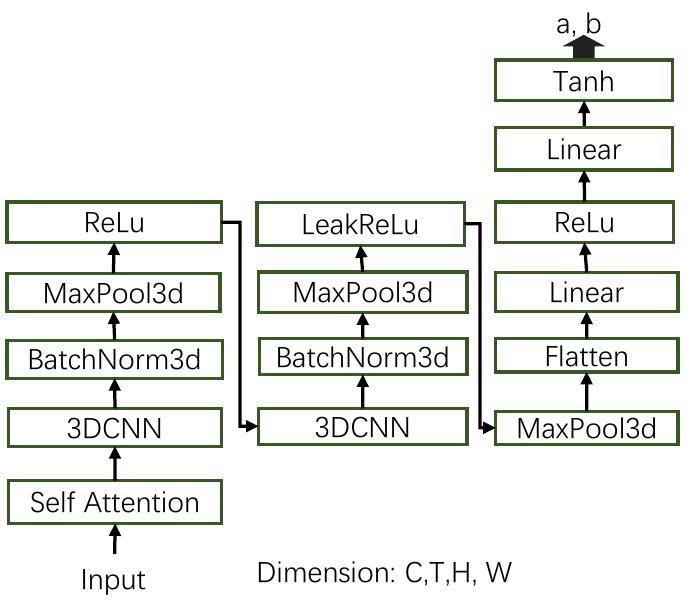}
    }
    \caption{The Structure of the Task-Specific Parameter Network and the Positioning Network. The Task-Specific parameter network uses all samples in an episode for a group of output $\lambda_1$ and $\lambda_2$, and a group of output $a$ and $b$ only maps to one sample in the Positioning network.
    }
 \label{fig:4_plus}
\end{figure*}

Linear sampling is used to align the action location duration by filtering the frames with less critical information or information that may mislead the video semantics. We use the first and second positioning networks to generate parameters for the Affine Transformation of the frames sequence in the time dimension. Because the probability distribution of videos in different tasks is often not the same, using a fixed network to generate the parameters of an Affine Transformation may lead to an inaccurate transformation. Therefore, we introduce a Task-Specific parameter network to create the modulation parameters for the first and second positioning networks according to the video samples in each episode. 
As shown in Figure \ref{fig:4}, the first and second positioning networks generate the affine parameters, which are used for sampling frames in the time axis. Meanwhile, the Task-Specific parameter network could generate modulation parameters for the two groups of affine parameters. In other words, modulation parameters will assign the degree of sampling from the two types of Affine Transformation. In the TSA module, our positioning network is designed based on 3D CNN, while our Task-Specific parameter network is designed based on 2D CNN. The detail of the structure for the Task-Specific parameter network and the positioning network is shown in Figure \ref{fig:4_plus}.
As in the formula \eqref{eq1} and formula \eqref{eq2}, $L_1$ is the first positioning network, and $L_2$ is the second positioning network. $f_X$ is the representation for both the support video and query video. In formula \eqref{eq3}, $T_{spec}$ is the Task-Specific parameter network that generates the modulation parameters for the two types of Affine Transformation, where the $F^s$ and $F^q$ are the samples of the support set and query set in each episode. 
It can be seen in formula \eqref{eq3} that a group of modulation parameters is generated using all the samples in each episode, so the modulation parameters are for each episode, but the affine parameters are just for a single video. $T_{\Delta}$ in formula \eqref{eq4} is the Affine Transformation for the location shift with the affine parameter $\Delta$. 
 \begin{equation} \label{eq1}
\rho_1 = [a_1, b_1]=L_1(f_X)
\end{equation}
 \begin{equation} \label{eq2}
 \rho_2=[a_2, b_2]=L_2(f_X)
\end{equation}
 \begin{equation} \label{eq3}
\Theta = [ \lambda_1, \lambda_2 ] = T_{spec}(F^s,F^q) 
\end{equation}
 \begin{equation} \label{eq4}
 \hat{f} = T_{\Delta}(f_X), \quad \Delta = \Theta \cdot 
 \begin{pmatrix}
 {\rho_1}\\
 {\rho_2}
 \end{pmatrix}
\end{equation}
We can refer to the paper ``Spatial Transformer Networks" \cite{stn-jaderberg2015spatial} for the details of this Affine Transformation. The difference is that the detail in the paper \cite{stn-jaderberg2015spatial} is for the spatial transformation of two-dimensional. Our work only focuses on the temporal transformation of one dimension. 
For more detail, we divide several interval coordinates from $-1$ to $1$ according to the number of frames. Then we use the positioning network $L_1$ and $L_2$ to generate the affine parameters, the $a_1$ and $b_1$, $a_2$ and $b_2$. Then, as the formula \eqref{eq4}, the Task-Specific Affine Transformation parameters are $a= \lambda_1 \times a_1 + \lambda_2 \times a_2$, $b= \lambda_1 \times b_1 + \lambda_2 \times b_2$, so the $grid\_coordinates = a \times interval\_coordinates+b$. We provide an input feature with several frames and a corresponding grid coordinate. Based on the coordinate information provided by each position in the grid, we fill the values of the corresponding position in the input to the position specified in the grid to obtain the final output.

\subsection{Multiple-level Transformer module(MLT)}
Our MLT module gets inspiration from TRX-related works \cite{perrett2021temporal-trx, thatipelli2022spatio-strm}. The difference is that our work focuses on the Multiple-level feature alignment in only one Transformer, and the TRX-related works use several Transformer instances, each instance using a single-level feature for alignment. Our MLT can obtain the correlation between features at different levels.
The detail is described as follows. 

In Figure \ref{fig:3}, following the rule of TSN\cite{wang2016temporal-tsn}, $M$ frames are sampled from each video. For simplicity, we use a 2-way 1-shot setting as an example, $f_{s1}$, $f_{s2}$ is support samples from the backbone, and $f_p$ is the query sample from backbone, the dimension of them is $C \times M \times H \times W$, where $C$ is the number of channels, $M$ is the number of frames, $H$ and $W$ are the dimension of 2D features. The $\hat{f_{s1}}, \hat{f_{s2}}, \hat{f_p}$ are the output of TSA, the dimension is the same as $f_{s1}$, $f_{s2}$ and $f_p$. The number of tuples for different cardinalities should be ``$C_M^1, C_M^2,\cdots, C_M^M$", where $C_i^M$ is the number of combinations selected for i frames in M frames. We use $w$ to indicate the cardinality of the tuples. Just as $w=1$ is for a single frame, which is for level one, $w=2$ is for a pair which is level two, $w=3$ is for a triple which is level three, $\cdots$, $w=m$ is for the features of level $m$. We generalise to possible tuples for any $w$, just as formula \eqref{eq5}:
\begin{equation}\label{eq5} 
\prod^w = {{(n_1, \cdots, n_w)}} \in N^w: \forall i  {(1 \le n_i < n_{i+1} \le  M)} 
\end{equation}
where the M is the number of frames we sampled.

The associated query representation with respect to the tuple with indices $p ={(p_1, \cdots, p_w) } \in \prod^w$, for the query tuple $p$, the query representation as formula \eqref{eq6}: 
 \begin{equation} \label{eq6} 
Q_p^w = [\Phi(q_{p_1}) + PE(p_1),\cdots,\Phi(q_{p_w}) + PE(p_w)] \in R^{w \times D}
\end{equation}
where $q_{p_1}, \cdots, q_{p_w}$ are the representation of frames with the frame index as $p_1, \cdots, p_w$. And $PE(\cdot)$ is a positional encoding given a frame index, and the $\Phi$ is a linear mapping which maps the dimension of each frame from $ R^{H \times W \times H}$ to $R^D$.

\textbf{For a tuple} based on different cardinality $w$, the linear  transformation as the formula \eqref{eq7} and formula \eqref{eq8}:
\begin{equation} \label{eq7} 
\Upsilon^w,\Gamma^w: R^{w \times D}\rightarrow R^{d_k},w\in \{1,\cdots m\}
\end{equation}  
\begin{equation} \label{eq8} 
\Lambda^w: R^{w \times D}\rightarrow R^{d_v},w\in \{1,\cdots m\}
\end{equation}
where $m \leq M $ is the largest cardinality we use.

In the TRX-related works, they have several Transformer instances according to how many cardinalities they use. If there are two cardinalities, as $\Omega=\{2,3\}$, there should be two instances of Transformer; the first one has the linear translation of $\Upsilon^2,\Gamma^2,\Lambda^2$, and the second one has the linear translation of $\Upsilon^3,\Gamma^3, \Lambda^3$. In our work, there is only one Transformer instance named Multiple-level Transformer that averages the information of several frames in each tuple of each cardinality into one, then all the linear translations as formula \eqref{eq7} and \eqref{eq8} could be the same as $\Upsilon^1,\Gamma^1,  \Lambda^1$. 

In our MLT, we define, when cardinality $w=1$, the number of the tuples is $N_1$, when $w=2$, the number of the tuples is $N_2$, ..., for $w=m$, the number of the tuples is $N_m$. For a sample, the feature dimension are $R^{N_1\cdot d_k}$ for $w=1$, $R^{N_2 \cdot d_k}$ for $w=2$, $\cdots$,  $R^{N_m \cdot d_k}$ for $w=m$. All the features belong to different cardinalities $w$ are contacted  together, the contact operation $\zeta$ is defined as formula \eqref{eq9}:
\begin{equation} \label{eq9}
\zeta:  R^{N_1 \times d_k}, R^{N_2 \times d_k}, \cdots, R^{N_m \times d_k} \rightarrow \\ 
        R^{(N_1+N_2+\cdots N_m) \times d_k}
\end{equation}
The more cardinalities (more levels) we use, the more tuples there will be. If we use the cardinalities from $w=1$ to $w=8$ in 8 frames, there will be 255 tuples. On one hand, a large number of tuples will take more memory. On the other hand, a large dimension of the feature will cause overfitting and affect the accuracy of the model. So, we generate a small number of tuple representations for the total tuples in a certain cardinality. The formula \eqref{eq9} should be modified as formula  \eqref{eq9'}:
\begin{equation} \label{eq9'}
\begin{split}
\hat{\zeta}:  
        &F^1_{net} (R^{N_1 \times d_k}),  F^1_{net} (R^{N_2 \times d_k}), \cdots, F^m_{net} (R^{N_m \times d_k}) 
        \rightarrow\\ 
        &R^{( \hat{N_1} + \hat{N_2}+ \cdots \hat{N_m} ) \times d_k}
\end{split}
\end{equation}
We define several small linear networks under cardinality $w$ as $F^w_{net}, w \in {\{1,2,\cdots,m\}} $, $N_w$ is the input dimension, which is equal to the total number of tuples under cardinality $w$, and $\hat{N_w}$ is the output dimension, which is a smaller number that we define.
For each cardinality, we use a linear network to get a small number of tuple representations that can represent the whole tuples. In this way, we avoid the problem of too many tuples under each cardinality, as we can still benefit from the variety of whole tuples under each cardinality.

We define the set for a small number of tuple representations created using the linear network from $\prod^w, w\in \{1,\cdots,m\}$ as $ \overline{\prod}^w$, $m$ is the largest cardinality we use. The correspondence between tuples of query $P$ and $S_{kt}^c$ which means tuple $t$ of support video $k$ in class $c$ is calculated as the formula \eqref{eq10}:
\begin{equation}  \label{eq10}
\begin{split}
a_{k,t,p}^c =  &L({\Gamma^w}\cdot S_{kt}^c)\cdot 
            L(\hat{\zeta}(\Upsilon^1 \cdot Q_p^1 , \Upsilon^2 \cdot Q_p^2, \cdots, \Upsilon^m \cdot Q_p^m)), \\
             &t \in  \overline{\prod} ^w, w \in \{1,\cdots,m\}
\end{split}
\end{equation}
where L is a standard layer normalisation \cite{ba2016layer-layer-norm} and $Q_p^1$,$Q_p^2$, $\cdots$, $Q_p^m$ is the feature of query $P$ that based on the cardinality $1$,$2$,$\cdots$,$m$. 

We apply the soft-max operation to acquire the attention map as the formula \eqref{eq11}. Because the attention operation is executed after the linear operation $\Upsilon$, and contact operation $\hat{\zeta}$. The across-attention matrix for every item includes not only the important measurement between single frames but also the important measurement for video semantics between different levels.
 \begin{equation} \label{eq11}
 \hat{a}_{k,t,p}^c = \frac{exp(a_{k,t,p}^c)/\sqrt{d_k}}{ \sum_{l,n}exp(a_{l,t,p}^c)/\sqrt{d_k}}
 \end{equation}
 Then, it is combined with value embedding of the support set as the following formula \eqref{eq12}:
 \begin{equation} \label{eq12}
 v_{kt}^c = {\Lambda}^w\cdot {S_{kt}^c}, w \in \{1,\cdots,m\} \quad and \quad  t \in  \overline{\prod}^w
 \end{equation}
 We compute the query-specific prototype with respect to the query $P$ as the following formula \eqref{eq13}:
\begin{equation} \label{eq13}
t_p^c =  \sum_{w=1}^{m}\sum_{k,t}  \hat{a}_{ktp}^c v_{kt}^c,  t \in  \overline{\prod}^w , w \in \{1,\cdots,m\}
 \end{equation}

\subsection{Optimal Transport distance}
The optimal Transport issue aims at finding optimal transportation with the lowest cost between two distributions. Assuming $a$ and $b$ are two distributions, $a \in R^n$ and $b \in R^m$, we aim to find a matrix $P \in R ^{n \times m}$, $P_{ij}$ means the probability of transferring from $a[i]$ to $b[j]$. We use $U(a,b)$ to represent all the possible solution space:
\begin{equation}  \label{eq14}
 U(a,b) = {\{P \in R^{n \times m} | P\bm{1}_m=a, P^T\bm{1}_n=b\}}
 \end{equation}
 Given a cot matrix $M \in R^{m \times n}$,  and the transfer plan $P$, the Optimal Transport problem is to find the most optimal transfer plan from $a$ to $b$, which can be expressed by the following formula\eqref{eq15}:
 \begin{equation}  \label{eq15}
 d_M(a,b) = \mathop{min}\limits_{P\in U(a,b)}\sum_{ij}P_{ij}M_{ij}
\end{equation}
 Using the Sinkhorn Algorithm \cite{cuturi2013sinkhorn}, the equation \eqref{eq15} can be optimized through iteration. We assume the optimal result of $U(r,c)$ is $P^*$. 
 
 The video matching problem could be described as an Optimal Transport problem\cite{lu2021few-cmot}. Using the Optimal Transport theory, we can force our model to pay attention to the semantics and appearance of the video. Our paper focuses on discrete Optimal Transport to formulate the video matching problem. After the MLT, we get the Multiple-level representations $f^{mul}$, including the single frame feature(level 1 feature) and other high-level features; the rank of level is the same as the number of frames in an origin tuple, as the formula \eqref{eq16}:  
\begin{equation}  \label{eq16}
\begin{split}
f^{mul} =  &\{ \underbrace{ f_1^1, f_1^2, \cdots, f_1^{l_1} }_{level_1}, \underbrace{ f_2^1, f_2^2, \cdots, f_2^{l_2} }_{level_2},\cdots \\
 &\underbrace{ f_m^1, f_m^2, \cdots, f_m^{l_m} }_{level_m}\} , (m \leq M )
\end{split}
\end{equation}
where $m$ is the largest cardinality we use, and $m \leq M $, $l_1$ is the number of frames we sample, $l_2$ is the number of small tuple representations created by the linear network for level $2$ tuples, and the $l_m$ is the number of small tuples created by the linear network for level $m$ tuples. After using the Transformer in the MLT module, according to Multiple-level features of a query $P$, which is represented as $\bar{f}_p$, 
and the Multiple-level features of support samples under the class $c$, we can get the query-specific Multiple-level prototype in support set for class $c$ as $t_p^c$. We assume the probability of $t_p^c$ and $\bar{f}_p$ is a uniform distribution. Then, we define an array named $\bm{level}$, the index of $\bm{level}$ represents the feature rank, and the value of each item in array $\bm{level}$ represents the number of tuple representations created by a linear network. The cost matrix $\mathbb{C}$ could be defined by the Euclidean distance (also other distances can be used here) between all the features at different levels as formula \eqref{eq17}.
\begin{equation}  \label{eq17}
\mathbb{C} = <\bar{f}_p, t_p^c >,\quad  \mathbb{C} \in R^{(\sum_{i=1}^{m}\mathbb{I}(i)level[i])\times(\sum_{j=1}^{m}\mathbb{I}(j)level[j])}
\end{equation}
\textbf{Note}, we perhaps only use parts of the levels, not always the whole levels, so we use an indicator function $\mathbb{I}$ to indicate whether we use the level.

In the end, the OT distance is defined as the formula \eqref{eq18}: 
\begin{equation}  \label{eq18}
dis_{ot}(\bar{f}_p, t_p^c) = \mathbb{C}P^*
\end{equation}
Using Optimal Transport distance to compare the Multiple-level features focuses on the semantic information alignment of the Multiple-level.

\subsection{Sequence mapping distance}
For the output of MLT, according to a certain query sample, we get the query-specific prototype of support set for class $c$ as $t_p^c$, and the representation of the query sample P as $\bar{f}_p$, the sequence mapping distance for each query video just as the formula \eqref{eq19}:
\begin{equation} \label{eq19}
   dis_{seq}(\bar{f}_p, t_p^c)=||\bar{f}_p, t_p^c||^2
\end{equation}

\subsection{Loss for the fusion of OT Distance and Sequence mapping Distance}
Given an N class problem, both the dimension of Optimal Transport distance and Sequence mapping distance should be $N$; we contact the distance element-wise as the formula \eqref{eq20}:
\begin{equation} \label{eq20}
   dis_{con}= contact(dis_{ot}, dis_{seq})
\end{equation}
where the dimension of $dis_{con}$ is $2N$.

Then we define a simple but useful fusion network as (1) A fully connected linear layer with the dimension of input is $2N$ that is the same as the dimension of $dis_{con}$, and output is $N$. (2) We use a Leaky ReLU operation after the linear mapping. (3)Then we add a Batch Normalization operation. 

We define the $dis_{fus}$ as the output of the fusion network. Probabilities over class $c \in {\{1,2,\cdots, N\}}$ that decide which class the query sample $P$ belongs to and the cross-entropy loss can be calculated as formula \eqref{eq21} and \eqref{eq22}:
\begin{equation}  \label{eq21}
p(y=c|P)=\frac{exp(dis_{fus}( \bar{f}_p, t_p^c ))}
{\sum_{\hat{c}=1}^{N} exp(dis_{fus}( \bar{f}_p, t_p^{\hat{c}} ))}
\end{equation}
\begin{equation} \label{eq22}
Loss_{fus} = -\frac{1}{||Q||} \sum_{(P,y^p) \in {Q,C}}Y_p\cdot log(p(y=c|P))
\end{equation}
where $Y_p \in \{0,1\}$ indicates if $y^p=c$. $Q$ and $C$ represent the query set and its corresponding class label set.

\section{Experiments}
\subsection{Datasets}
We train and evaluate the model on four datasets commonly used in the field of few-shot action recognition, including UCF101\cite{soomro2012ucf101}, HMDB51\cite{kuehne2011hmdb}, Kinetics400\cite{carreira2017quo-kinetics} and SSV2\cite{goyal2017something}. Because our work focuses on few-shot learning, we need to segment the dataset properly. For Ssv2 and Kinesics, we use the split method of CMN\cite{zhu2018compound}, CMN-J\cite{zhu2020label}. The split method randomly selects a mini-dataset containing 100 classes, including 64 classes for training, 12 classes for validation, and 24 classes for testing, of which each class contains 100 samples. For Ssv2, we also use the OTAM\cite{cao2020few-otam} split method, which is similar to the Ssv2-CMN split method, except it uses all samples inside the class. In the OTAM, it contains $77500/1926/2854$ videos included for $train/val/test$ respectively, as some videos can not be found now, so in our work, it contains $67013/1926/2854$ videos for $train/val/test$ respectively. The UCF101 contains 101 action classes, according to the split in ARN\cite{zhang2020few-ARN}, we selected 70 classes for training, 10 classes for verification, and 21 classes for testing; it contains $9154/1421/2745$ videos for $train/val/test$ respectively. For the HMDB51, we also follow the split rule of ARN \cite{zhang2020few-ARN}, 31 classes for training, 10 classes for verification, and 10 classes for testing, with $4280/1194/1292$ videos for $train/val/test$.
\subsection{Implementation Details}
Following the TSN\cite{wang2016temporal-tsn} and previous methods, we sparsely and uniformly sample several ${(i.e. T=8)}$ frames for each video. We flip each frame horizontally and randomly crop the center region of 224 × 224 for the training data augmentation. For the backbone, we use ResNet-50\cite{he2016deep-resnet}, which is pre-trained on ImageNet\cite{deng2009imagenet}. We remove the final fully connected layer and get the raw frame-level feature with 2048 dimensions. In the meta-training phase, for the Ssv2 dataset, we randomly sample the support set and training set and give 100,000 episodes; for the Kinetics, we sample 6,000 episodes; and for HMDB51 and UCF101, we also sample 6,000 episodes. In the meta-testing phase, we sample 10,000 episodes and get the average accuracy of each episode. We utilize the SGD optimizer with a learning rate of 0.001 and 0.0001, and we train our model on Nvidia 3090 GPU. During training, we thus average gradients and backpropagate once every 16 iterations. For testing, we use only the center crop to augment the video. 

\subsection{Comparison with previous works}
We compare the performance of our TSA-MLT with lots of state-of-the-art methods. Table \ref{table:1} shows the comparison with various state-of-the-art methods on UCF101 and HMDB51. Table \ref{table:2} shows the comparison of Kinetics and Ssv2. The standard 5-way 1-shot setting and 5-way 5-shot setting tasks are scheduled for the test. 

In Table \ref{table:1}, for TRX and STRM, in the HDMB51 and UCF101, we give our implementation results for the 1-shot setting marked as red, as there is no result for the 1-shot setting. Based on the HDMB51, for the 5-shot setting, the performance of TSA-MTL has been improved to a certain extent compared with previous methods, and the accuracy has reached $78.2\%$. For the 1-shot setting, TSA-MLT accuracy is $57.9\%$, $3.8\%$ higher than STRM, and $2\%$ higher than TRX. For UCF101, 5-shot setting, the accuracy of TSA-MTL is $97.1\%$, which is $0.2\%$ higher than STRM and $1\%$ higher than TRX. For the 1-shot setting, the accuracy of TSA-MLT is $80.6\%$. \textbf{For 5-shot setting based on HMDB51, and the 5-shot setting based on UCF101 we achieve the state-of-the-art result.}

In Table \ref{table:2}, for TRX and STRM, in the Kinetics and Ssv2, we mark our implementation results as red; the results are a little lower than that published, which may be due to the machine environment or random seed. In Kinetics, for the 5-shot setting, the accuracy published for TRX is $85.9\%$, our implementation is $85.1\%$, the accuracy published for STRM is $86.7\%$, our implementation is $85.9\%$. The accuracy of TSA-MLT is $87.1\%$, and it is higher than our implementation for both TRX and STRM($2\%$ and $1.2\%$ improved), but only $1.2\%$ higher than that published by TRX and $0.4\%$ higher than that published by STRM. For the 1-shot setting, TRX releases $63.6\%$, and our implementation is $63.4\%$. STRM does not release the result for the 1-shot setting, and our implementation is $65.3\%$. The testing result of the 1-shot setting for TSA-MLT is $66.8\%$. \textbf{For a 5-shot setting based on Kinetis, 
the state-of-the-art result is 87.4\%. Here, we should pay attention that the work of TADRNet follows the rule of multi-modality, but the work model is mono-modality, so the result of our TSA- MLT 87.1\% is also competitive.} 

In Ssv2-all, for the 5-shot setting, TRX publishes the accuracy as $64.6\%$, our implementation is $63\%$, STRM publishes the accuracy as $68.1\%$, our implementation is $64.8\%$, and for TSA-MLT, the accuracy is $65.1\%$. For the 1-shot setting, the result published by TRX is $42\%$. STRM also does not publish for a 1-shot setting, while our implementation is 42.9\%. The accuracy of TSA-MLT is $43.8\%$. Based on Ssv2-all, our results for 1-shot and 5-shot settings are higher than both the TRX and STRM of our implementation and are close to the state-of-the-art. In SSV2-part, for the 5-shot setting, TRX publishes the result as $59.4\%$, our implementation of STRM is $58.6\%$, a little lower than TRX, and our TSA-MLT is $60.9\%$, $1.5\%$ higher than TRX. For the 1-shot setting, TRX publishes the result of $36.0\%$, for STRM, our implementation is $38.3\%$, and the accuracy of TSA-MLT is $37.0\%$, $1\%$ higher than TRX, and $1.3\%$ lower than STRM. \textbf{For 5-shot setting based on Ssv2-part, our TSA-MLT achieve the state-of-the-art result $60.9\%$.} The other results are also competitive.

\begin{table*}[htbp]
  \centering
  \caption{Comparison on UCF101 and HMDB51. Classification accuracy results for 5-way 1-shot setting and 5-way 5-shot setting experiments are shown. $\star$ means the item contains our implementation. The data with * means the state of the art. The data marked in red indicates the results of our implementation. This table misses Matching Net, CMN, TARN, and SMMF compared to Table \eqref{table:2} because there are no corresponding results in their works.}
  \label{table:1}
  \scalebox{1}{
  \begin{tabular}{l|l|l|cc|cc}
    \toprule 
    \multicolumn{1}{c|}{\multirow{2}[1]{*}{Method}} & \multicolumn{1}{c|}{\multirow{2}[1]{*}{Reference}} & \multicolumn{1}{c|}{\multirow{2}[1]{*}{Backbone}} & \multicolumn{2}{c|}{HMDB51} &\multicolumn{2}{c}{UCF101} \\
      &    &       & 1-shot & 5-shot & 1-shot & 5-shot \\ \hline
    ProtoNet\cite{snell2017prototypical_few_shot_learning2} & NeurIPS’2017 & ResNet-50 &54.2    &68.4      & 74.0      &89.6  \\
    ARN\cite{zhang2020few-ARN} & ECCV’2020 &C3D&44.6&59.1&62.1& 84.8 \\
    AMeFu-Net\cite{fu2020depth} & MM’2020 &ResNet-50&60.2&75.5&85.1& 95.5 \\
    PAL\cite{zhu2021few-PAL}  & ARXIV'2021 &ResNet-50&60.9&75.8&85.3&95.2 \\
    OTAM\cite{cao2020few-otam}  & CVPR’2020&ResNet-50&54.5&66.1&79.9& 88.9 \\
    CMOT\cite{lu2021few-cmot}  & ARXIV'2021 &C3D& \textbf{$64.6^*$}  &77.0 & \textbf{$90.4^*$} & 95.7 \\
    HyRSM\cite{wang2022-hybrid}& CVPR’2022 &ResNet-50&60.3&76.0&83.9&94.7 \\
    MTFAN\cite{wu2022motion}& CVPR’2022 &ResNet-50&59.0&74.6&84.8&95.1 \\
    TA2N\cite{li2022ta2n}  & AAAI’2022&ResNet-50&59.7& 73.9&81.9& 95.1\\
    MPRE\cite{liu2022multidimensionalMPRE} & TCSVT'2022 &ResNet-50&57.3& 76.8&82.0& 96.4\\
    TADRNet\cite{2023Task-AwareDual-Representation} & TCSVT'2023 &ResNet-50&64.3& \textbf{$78.2^*$} &86.7& 96.4\\
    MoLo\cite{wang2023molo} & CVPR'2023 &ResNet-50&60.8& 77.4&86.0& 95.5\\
    HyRSM++ \cite{wang2023hyrsm++} & ARXIV'2023 &ResNet-50&61.5& 76.4&85.8& 95.9\\
    $\star$TRX \cite{perrett2021temporal-trx} & CVPR’2021 &ResNet-50& \textcolor{red}{55.9}  &75.6&\textcolor{red}{77.3}&96.1 \\
    $\star$STRM \cite{thatipelli2022spatio-strm} & CVPR’2022 &ResNet-50 & \textcolor{red}{54.1}&77.3&\textcolor{red}{79.2}&96.9 \\
    \midrule
    Ours(TSA-MLT) & &ResNet-50&\textcolor{red}{57.9}& \textcolor{red}{$78.2^*$}  &\textcolor{red}{80.6}&\textcolor{red}{$97.1^*$} \\
    \bottomrule 
    \end{tabular}%
    }
\end{table*}%

\begin{table*}[htbp] 
  \centering
  \caption{Comparison on Kinetics and Ssv2 datasets, for the classification accuracy. Results of 5-way 1-shot and 5-way 5-shot experiments are shown. The Ssv2-part follows the split from \cite{zhu2018compound}, and the Ssv2-all follows the split from \cite{cao2020few-otam}. $\star$ means the item contains our implementation. The data marked as red in the parenthesis indicates the results of our implementation, reported results shown outside the parenthesis. The data with * means the state of the art.}
    \label{table:2}
    \scalebox{1}{
    \begin{tabular}{l|l|l|cc|cc|cc}
    \toprule
    \multicolumn{1}{c|}{\multirow{2}[1]{*}{Method}} & \multicolumn{1}{c|}{\multirow{2}[1]{*}{Reference}} & \multicolumn{1}{c|}{\multirow{2}[1]{*}{Backbone}} & \multicolumn{2}{c|}{Kinetics} & \multicolumn{2}{c|}{Ssv2-part} & \multicolumn{2}{c}{Ssv2-all} \\
      &    &       & 1-shot & 5-shot & 1-shot & 5-shot & 1-shot & 5-shot \\\hline
    Matching Net\cite{vinyals2016matching—few_shot_learning5}& NeurIPS’2016 &ResNet-50 & 53.3  &74.6&34.4&43.8&-& - \\
    ProtoNet \cite{snell2017prototypical_few_shot_learning2}& NeurIPS’2017&ResNet-50&64.5   &77.9&33.6&43.0&-&-\\
    CMN\cite{zhu2018compound} & ECCV’2018 &ResNet-50&60.5&78.9&36.2&48.8&-&- \\
    ARN\cite{zhang2020few-ARN} & ECCV’2020 &C3D & 63.7&82.4&-&-&-&- \\
    AMeFu-Net\cite{fu2020depth} & MM’2020 &ResNet-50 & \textbf{$74.1^*$}&86.8&-&-&-&- \\
    SMNF\cite{qi2020few-smfn} & MM’2020 &ResNet-50 & 63.7&83.1&-&-&-&- \\
    PAL\cite{zhu2021few-PAL} & ARXIV'2021 &ResNet-50&\textbf{$74.2^*$}&\textbf{$87.1^*$}&-&-&46.4 &62.6  \\
    OTAM\cite{cao2020few-otam} & CVPR'2020 &ResNet-50&73.0&85.8&-&-&42.8&52.3 \\
    CMOT\cite{lu2021few-cmot}  & ARXIV'2021 &C3D&-&-&-&-&46.8& 55.9  \\
    HyRSM\cite{wang2022-hybrid} & CVPR’2022 &ResNet-50&73.7&86.1& 40.6 &56.1&54.3& 69.0 \\
    MTFAN\cite{wu2022motion} & CVPR’2022 &ResNet-50&74.6&\textbf{$87.4^*$}&-&-& 45.7 & 60.4 \\
    TA2N\cite{li2022ta2n} & AAAI’2022 &ResNet-50&73.0&85.8&-&-& \textbf{$47.6^*$}& 61.0 \\
    MPRE\cite{liu2022multidimensionalMPRE} & TCSVT'2022 &ResNet-50&70.2&85.3&-&-& 42.1& 58.6 \\
    TADRNet\cite{2023Task-AwareDual-Representation} & TCSVT'2023 &ResNet-50&75.6&\textbf{$87.4^*$}&-&-& 43.0& 61.1 \\
    MoLo\cite{wang2023molo} & CVPR'2023 &ResNet-50&74.0&85.6&\textbf{$42.7^*$}&56.4& \textbf{$56.6^*$} & \textbf{$70.6^*$} \\
    HyRSM++ \cite{wang2023hyrsm++} & ARXIV'2023 &ResNet-50&74.0&86.4&42.8&58.0&55.0& 69.8\\
    $\star$TRX\cite{perrett2021temporal-trx} &  CVPR’2021 &ResNet-50&63.6(\textcolor{red}{63.4})&85.9(\textcolor{red}{85.1})&36.0&59.4&42.0&64.6(\textcolor{red}{63.0}) \\
    $\star$STRM\cite{thatipelli2022spatio-strm}&  CVPR’2022 & ResNet-50&\textcolor{red}{65.3}&86.7 (\textcolor{red}{85.9}) &\textcolor{red}{38.3}&\textcolor{red}{58.6}  &\textcolor{red}{42.9}& 68.1(\textcolor{red}{64.8}) \\
    \midrule
    Ours(TSA-MLT) & &ResNet-50&\textcolor{red}{66.8}&\textcolor{red}{$87.1$}  &\textcolor{red}{37.0}&\textcolor{red}{$60.9^*$}&\textcolor{red}{43.8}& \textcolor{red}{65.1} \\
    \bottomrule
    \end{tabular}%
    }
\end{table*}%

\subsection{Ablation Study}
\subsubsection{Analysis of Model Components}

In this part, we design and conduct some comparative experiments on Kinetics and UCF101 to evaluate the effectiveness of each part we construct. Here, we still use the 5-shot and 1-shot for the evaluation. 

In Table \ref{table:3}, the experiment demonstrates that we have two critical sub-modules, and the $fusion\_loss$ is for final classification. Except for the ablation of sub-modules, we still want to verify the fusion of Optimal Transport Distance and Sequence mapping Distance is effective, so we define $sequence\_loss$ and $ot\_loss$ which are similar to the formula \eqref{eq21} and \eqref{eq22}, the only difference is that $dis_{seq}$ as formula \eqref{eq19} is used for $sequence\_loss$ and $dis_{ot}$ as formula \eqref{eq18} is used for $ot\_loss$. In the experiment, our ablation parts are TSA, MLT, $sequence\_loss$, $ot\_loss$, and $fusion\_loss$.

Overall, as the accuracies in Table \ref{table:3}, using TSA and MLT with the fusion loss, the maximum performance can be achieved. We can summarize the ablation experiment:
(1) Only using the MLT, our model can get a better performance than TRX under our implementation. it can be refer in Table \ref{table:1} and \ref{table:2}.
(2) Under 5-shot setting, using the $fus\_loss$ which is computed by the $dis_{fus}$, the accuracy is increased compared to both $dis_{ot}$ and $dis_{seq}$, because the $dis_{ot}$ just brings the supplementary for semantic information and $dis_{seq}$ just brings the order mapping information, and the $dis_{fus}$ carries both.
(3) Under the 1-shot setting, when using the $ot\_loss$ which is computed by the $dis_{ot}$, the accuracy is a little higher than using  $dis_{fus}$, perhaps because when there is only one support sample, the information is not enough to summarize the sequence distribution. So the lower accuracy of the $dis_{seq}$ perhaps affects the accuracy of $dis_{fus}$.
(4) The accuracy in most items with the TSA is higher than without it can approve the effect of the TSA. However, for HMDB51, the improvement is small. Because HMDB51 focuses on spatial and appearance, the probability of the phenomenon that several sampled frames contain insignificant information is not high. The Kinetics also focuses on spatial information, but it is more complex than HMDB51, and the Ssv2 focuses on temporal information. So, the improvement of these two datasets is a little higher than HMDB51.

\begin{table*}[htbp]
  \centering
  \caption{Ablation for the Analysis of Model Components. The experiment is for the module of TSA and MLT with three types of loss. The Ablation study is under the datasets of Kinetics, HMDB51 and Ssv2-all.}
  \scalebox{1}{
    \begin{tabular}{c|c|c|c|c|c|c|c|c|c|c|c}
    \toprule
    \multicolumn{1}{c|}{\multirow{2}[2]{*}{method}} 
    & \multicolumn{2}{c|}{module} 
    & \multicolumn{3}{c|}{loss} 
    & \multicolumn{2}{c|}{Kinetics} 
    & \multicolumn{2}{c|}{HMDB51} 
    & \multicolumn{2}{c}{Ssv2-all} \\
          & \multicolumn{1}{l}{MLT} 
          & \multicolumn{1}{l|}{TAS} 
          & \multicolumn{1}{l}{sequence\_loss} 
          & \multicolumn{1}{l}{ot\_loss} 
          & \multicolumn{1}{l|}{fusion\_loss} 
          & \multicolumn{1}{l}{1-shot} & \multicolumn{1}{l|}{5-shot}
          & \multicolumn{1}{l}{1-shot} & \multicolumn{1}{l|}{5-shot}
          & \multicolumn{1}{l}{1-shot} & \multicolumn{1}{l}{5-shot}\\
    \midrule
    \multicolumn{1}{c|} {(1)} &\ding{51}&\ding{55}&\ding{51}&\ding{55}&\ding{55}&64.2&85.4&52.9&75.8&41.9&63.0\\
    \multicolumn{1}{c|} {(2)} &\ding{51}&\ding{55}&\ding{55}&\ding{51}&\ding{55}&70.5&85.5&59.7&77.0&42.2&62.2\\
    \multicolumn{1}{c|} {(3)} &\ding{51}&\ding{55}&\ding{55}&\ding{55}&\ding{51}&66.5&86.5&57.5&78.1&42.0&63.9\\
    \multicolumn{1}{c|} {(4)} &\ding{51}&\ding{51}&\ding{51}&\ding{55}&\ding{55}&66.7&86.0&57.3&76.9&43.2&64.4\\
    \multicolumn{1}{c|} {(5)} &\ding{51}&\ding{51}&\ding{55}&\ding{51}&\ding{55}&71.0&85.5&60.0&77.1&43.8&64.4\\
    \multicolumn{1}{c|} {(6)} &\ding{51}&\ding{51}&\ding{55}&\ding{55}&\ding{51}&66.8&87.1&57.9&78.2&43.8&65.1\\
    \bottomrule
    \end{tabular}%
    }
  \label{table:3}%
\end{table*}%

\begin{table*}[htbp]
  \centering
  \caption{Compare the accuracy of TLA-MLT with varying cardinalities.} \label{table:4}
  \scalebox{1}{
    \begin{tabular}{l|l|l|c|c}
    \toprule
    No. & Cardinalities & tuples number & Kinetics & HMDB51 \\
    \midrule
   (1) & $\Omega$ = \{1\}   & 8       &  85.3   & 75.8 \\
   (2) & $\Omega$ = \{1,2,3\} & 8+4+3(15)  & 85.6   &77.9  \\
   (3) & $\Omega$ = \{1,2,4\} & 8+4+2(14) & 86.5   & 77.8 \\
   (4) & $\Omega$ = \{1,2,5\} & 8+4+1(13)  & 86.0   & 76.7 \\
   (5) & $\Omega$ = \{1,3,4\} & 8+3+2(13) & 85.3   & 78.2 \\
   (6) & $\Omega$ = \{1,3,5\} & 8+3+1(12)  & 84.9    & 76.9 \\
   (7) & $\Omega$ = \{1,4,5\} & 8+2+1(11)  & 85.2 & 75.9 \\
   (8) & $\Omega$ = \{1,2,3,4\} & 8+4+3+2(17)  &87.1 & 78.2 \\
   (9) & $\Omega$ = \{1,2,3,5\} & 8+4+3+1(16)  & 85.9     &76.4  \\
   (10) & $\Omega$ = \{1,2,4,5\} & 8+4+2+1(15)    & 86.6     & 77.7 \\
   (11) & $\Omega$ = \{1,3,4,5\} & 8+3+2+1(14)   & 85.8      & 78.0\\
   (12) & $\Omega$ = \{1,2,3,4,5\} & 8+4+3+2+1(18) & 85.4   & 76.2 \\
    \bottomrule
    \end{tabular}%
    }
\end{table*}%

\subsubsection{Compare with varying cardinalities}

 In Table \ref{table:4}, using $TSA+MLT+fusion\_loss$, several different combinations of the cardinalities are selected, and it shows the accuracy of Kinetics and HMDB51 under a 5-shot setting. There are several linear networks as $F^w_{network}, w \in \{1,2,\cdots,m\}$, as the formula \eqref{eq9'}, $N_w$ as the dimension of input is the number of all tuples under specific cardinality $w$ and $\hat{N}_w$ is equal to the number of tuple representations we will use. We just use five cardinalities in our experiment, for cardinality $w=1$, we define $\hat{N_1} = 8$, for cardinality $w=2$ we define $\hat{N_2} = 4$, for cardinality $w=3$ we define $\hat{N_3} = 3$ for cardinality $w=4$ we define $\hat{N_4} = 2$, for cardinality $w=5$ we define $\hat{N_5} = 1$. Among all the combinations in the table, we find that the $w=\{1,2,3,4\}$ with the number of tuple representations as $\{8,4,3,2\}$ is the optimal choice. It is also the configuration for Table \ref{table:1}, Table \ref{table:2} and Table \ref{table:3}. For TRX, the optimal choice is $w=\{2,3\}$, and the number of tuples is $84(28+56)$. For STRM, the optimal choice is based on the $w= \{2\}$, and the number of tuples is $28$. We can see the total number of tuples for our model is $17$ under the optimal choice, much less than the other methods.

\subsubsection{Performance of our model under the 5-way k-shot setting}

\begin{figure*}[htbp]
    \centering
    \subfloat[Accuracy for UCF101]{
    \label{fig:5.a}
    \includegraphics[width=8cm]{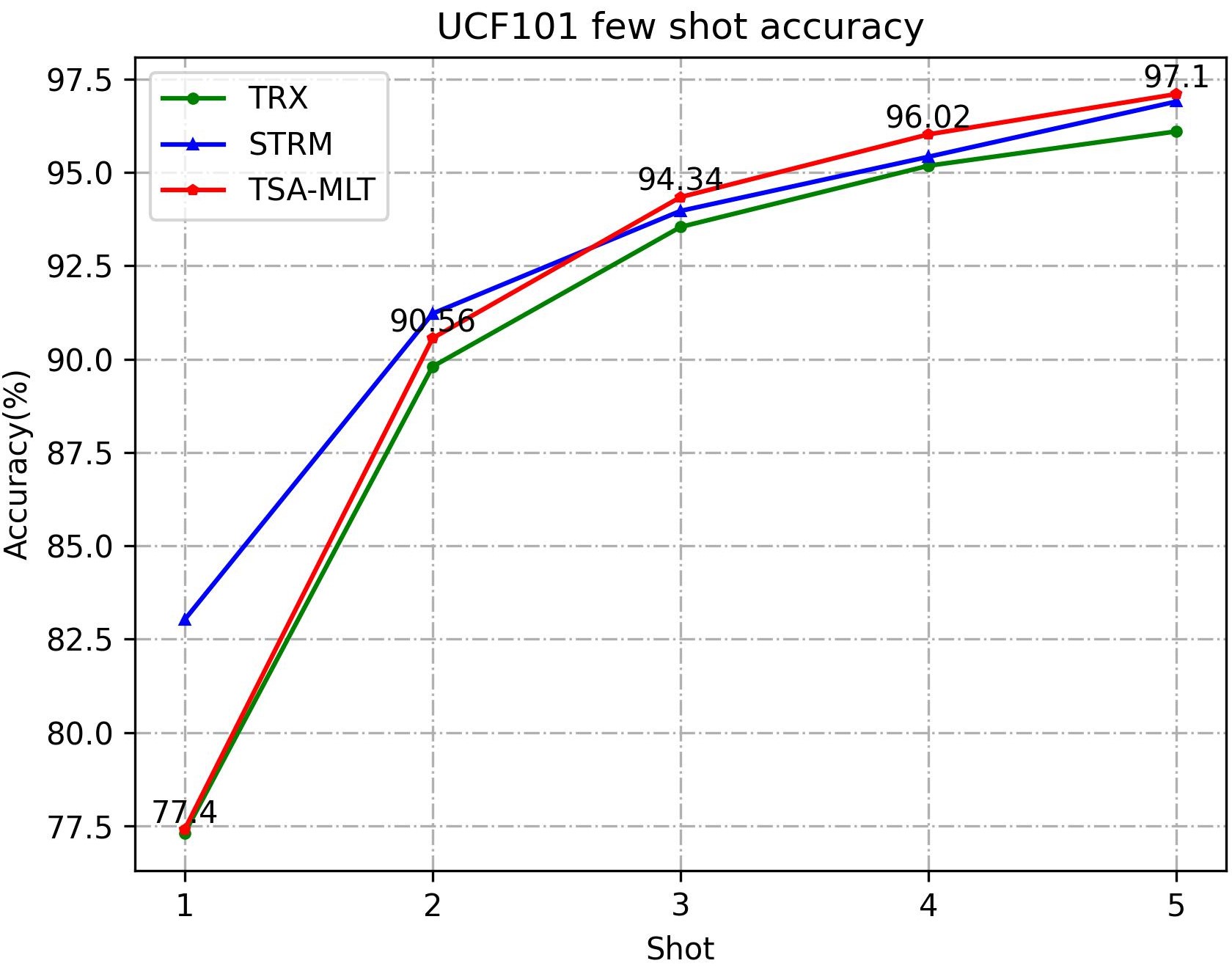}
    }
    \subfloat[Accuracy for SSV2-all]{
    \label{fig:5.b}
    \includegraphics[width=8cm]{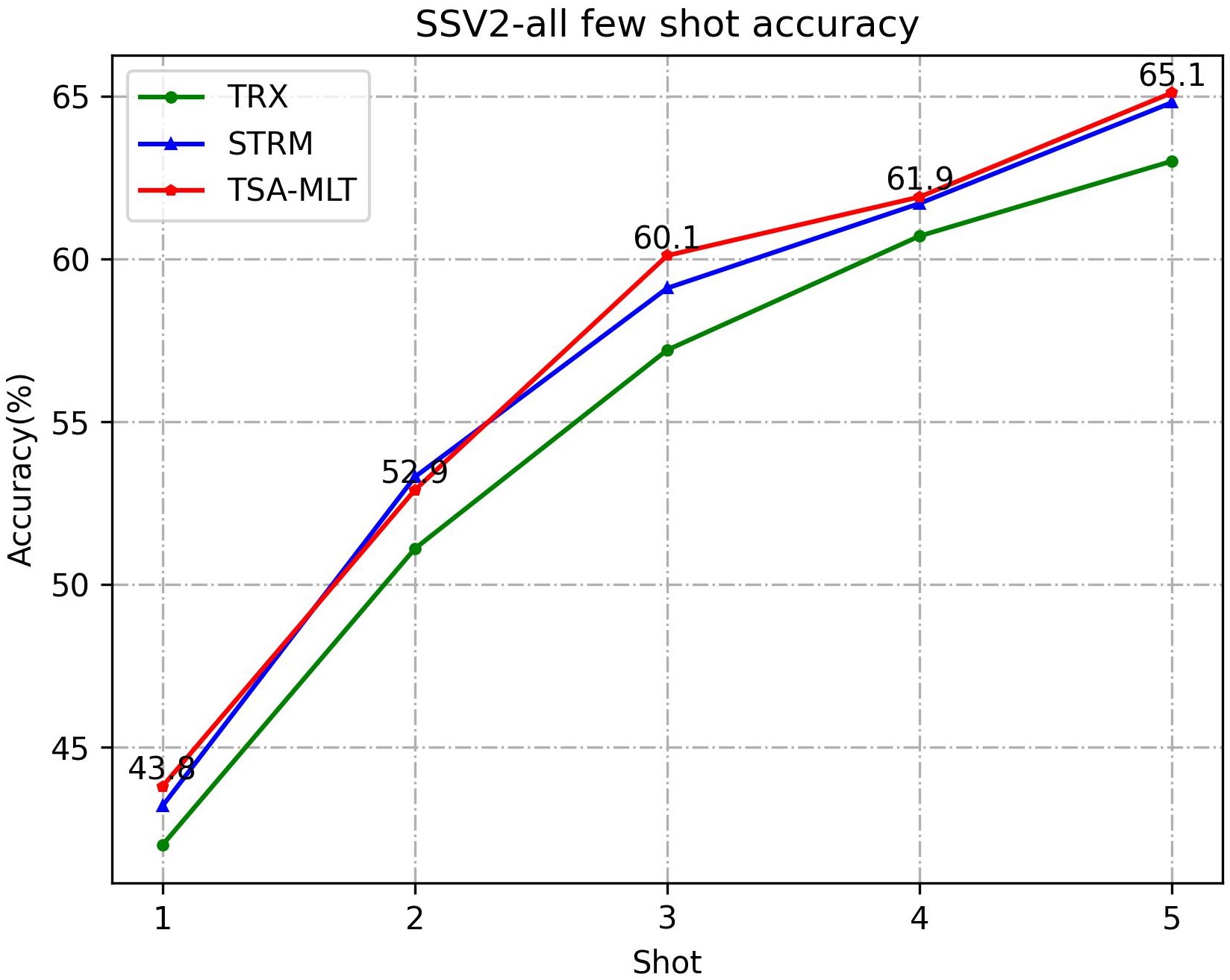}
    }
    \caption{The curve of the accuracies from 1-shot to 5-shot setting for the dataset of UCF101 and Ssv2-all.
    }
\label{fig:5}
\end{figure*}

We test the performance of the model on Ssv2-all and UCF10. Figure \ref{fig:5} shows the accuracy for 1-shot to 5-shot of UCF101 and Ssv2-all. While the shots increase, the accuracy of each model also improves. It is worth noting that our method has consistently maintained leading performance compared to TRX and STRM models, except for UCF101 in 1-shot and 2-shot; the accuracy of our model is higher than others, which shows the robustness of our method. In Figure \ref{fig:5}, the accuracies follow our implementation.

\section{Visualization}
\subsection{Accuracy comparison for different class}
\begin{figure}
\centering
\includegraphics[width=9cm]{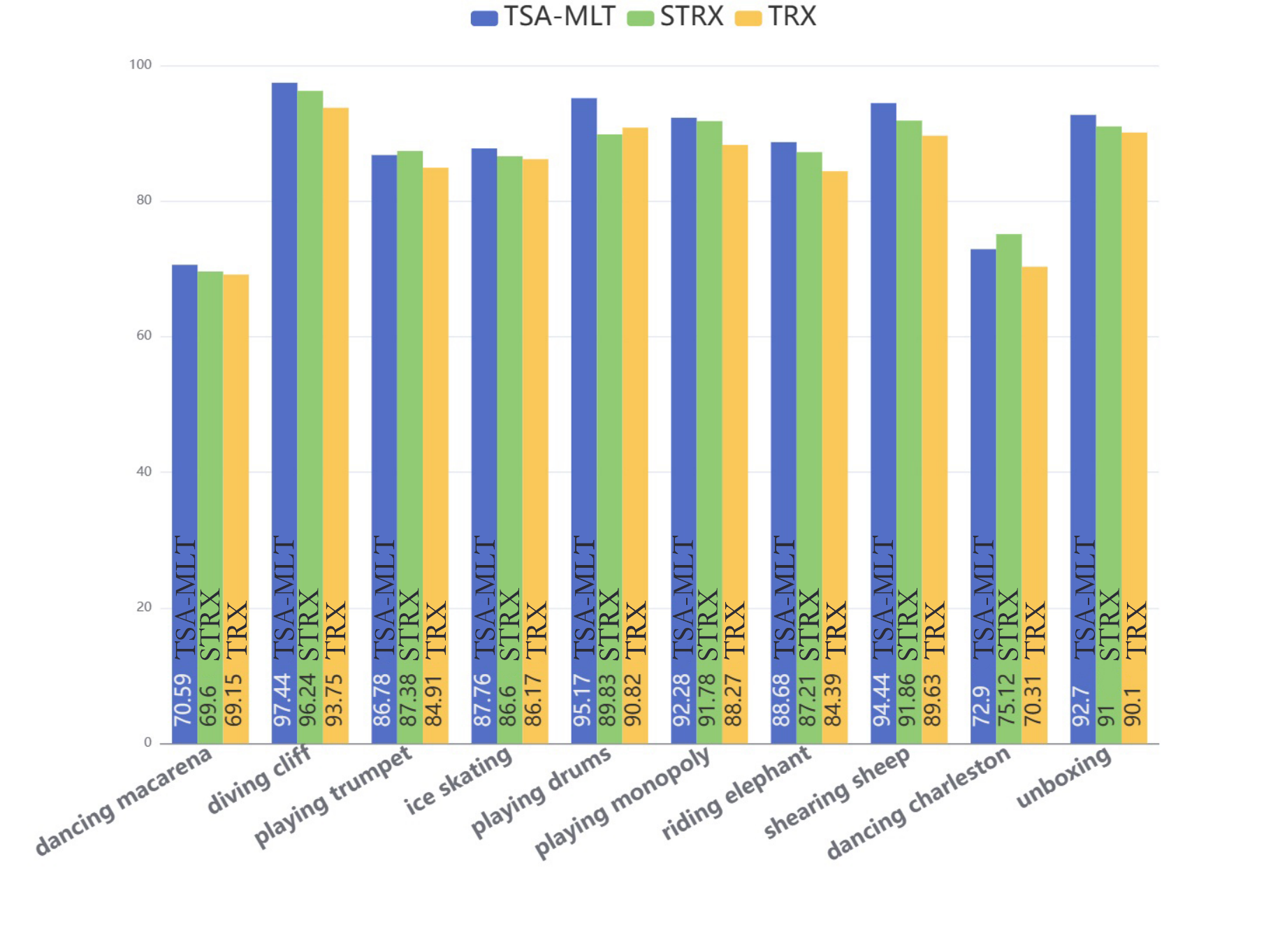}
\caption{
The comparison of TRX, STRM, and TSA-MLT for the accuracy of certain action classes in the Kinetics dataset.}
\label{fig:6}
\end{figure}

Compare with the TRX and STRM for the accuracy of the Kinetics. In our paper, we select 10 classes and get the accuracy under the 5-shot setting. In Figure \ref{fig:6}, it can be seen the accuracy of our model is improved for most of the classes from TRX and STRM except for the class ``dancing charleston". For the class ``playing drums" the accuracy has the most improvement, from the accuracy of $90.82\%$ of TRX and the accuracy of $89.83\%$ of STRM to the accuracy of $95.17\%$ of our model. The average accuracy for these 10 action classes is improved from $84.75\%$ of TRX  and $86.66\%$ of STRM to $87.87\%$ of our TSA-MLT.

\subsection{Distribution comparison}
\begin{figure}[htbp]
    \centering
    \subfloat[TRX prototype]{
    \label{fig:7.a}
    \includegraphics[width=0.16\textwidth]{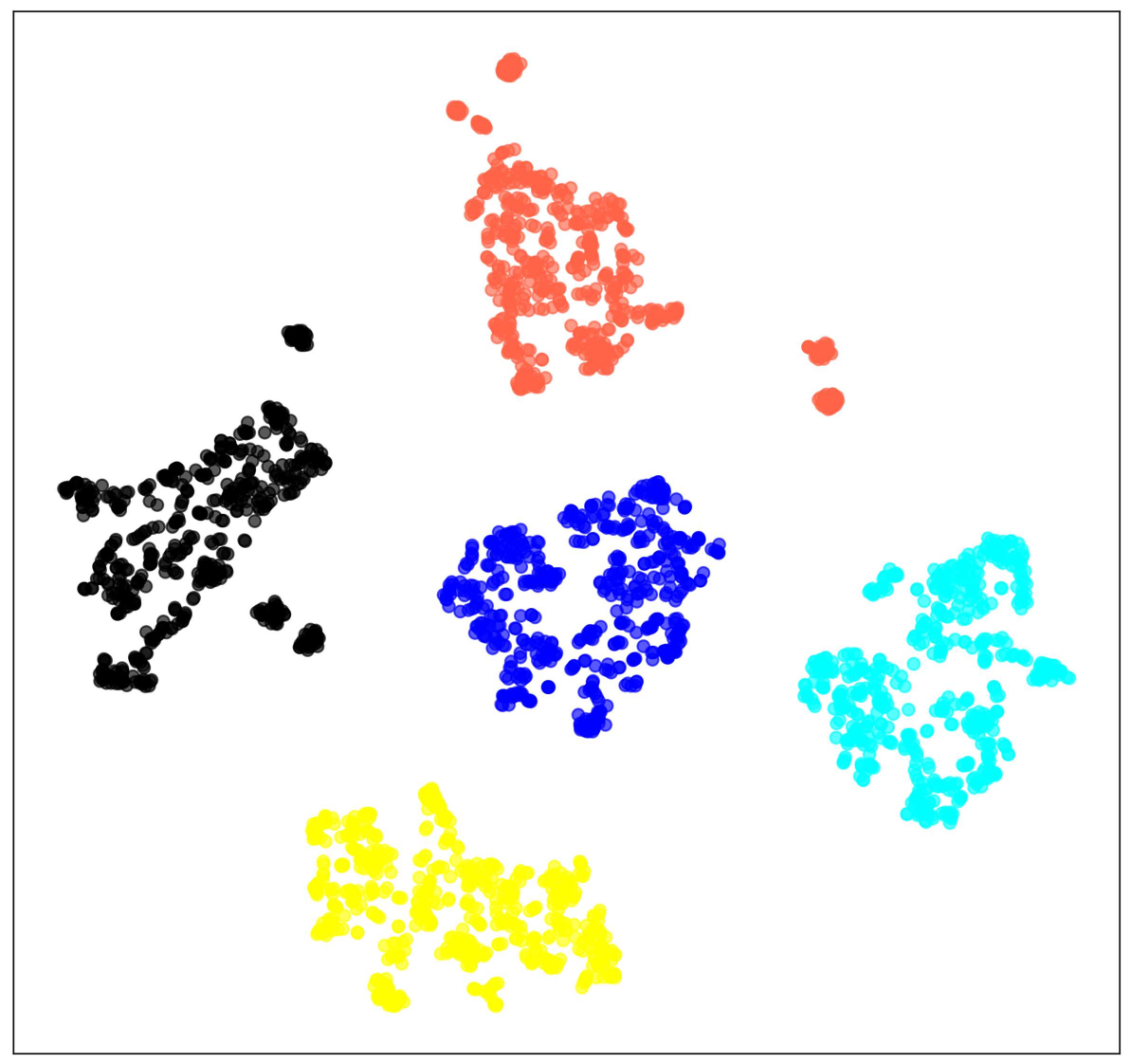}
    }
    \subfloat[STRM prototype]{
    \label{fig:7.b}
    \includegraphics[width=0.16\textwidth]{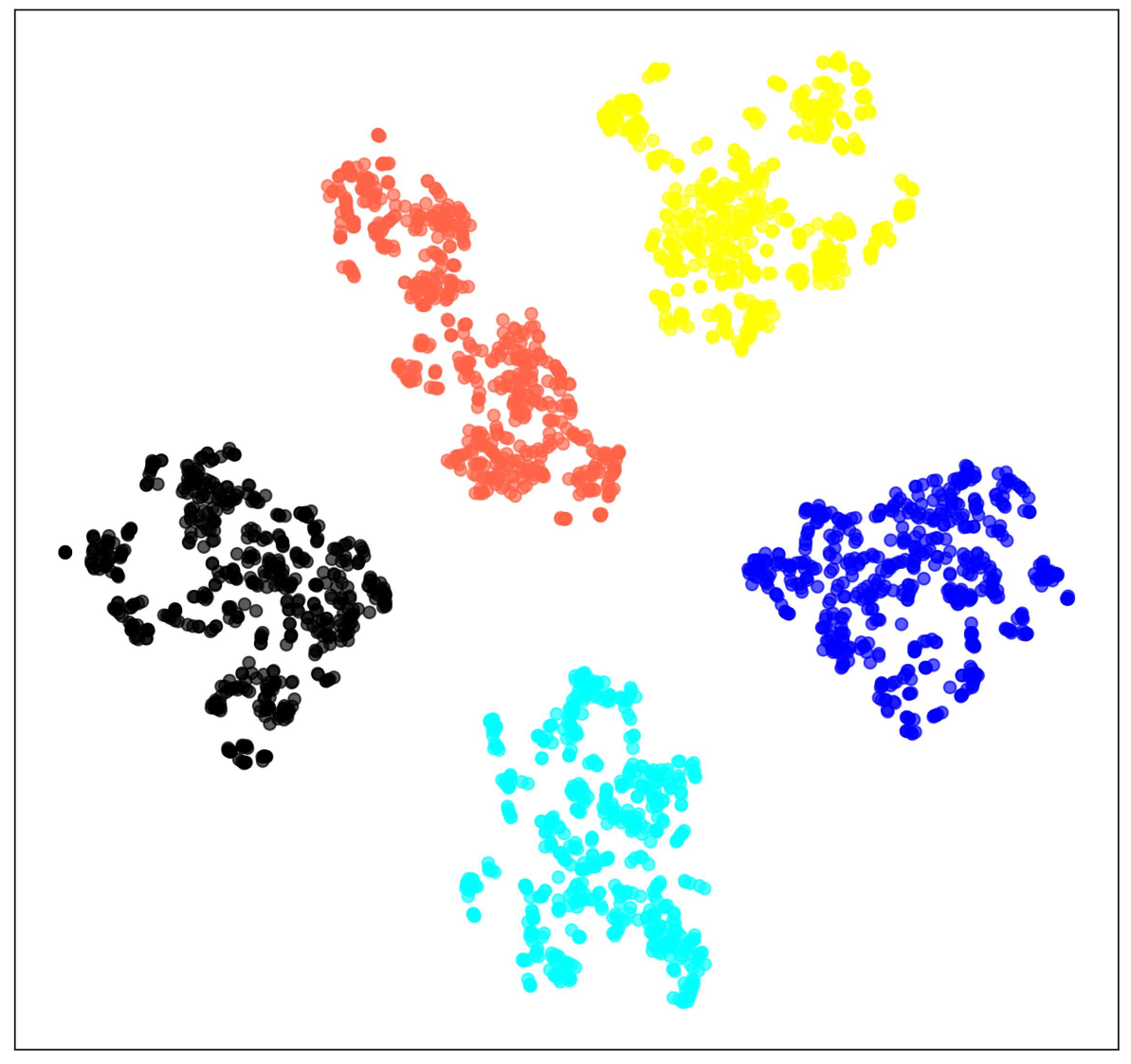}
    }
    \subfloat[TSA-MLT prototype]{
    \label{fig:7.c}
    \includegraphics[width=0.16\textwidth]{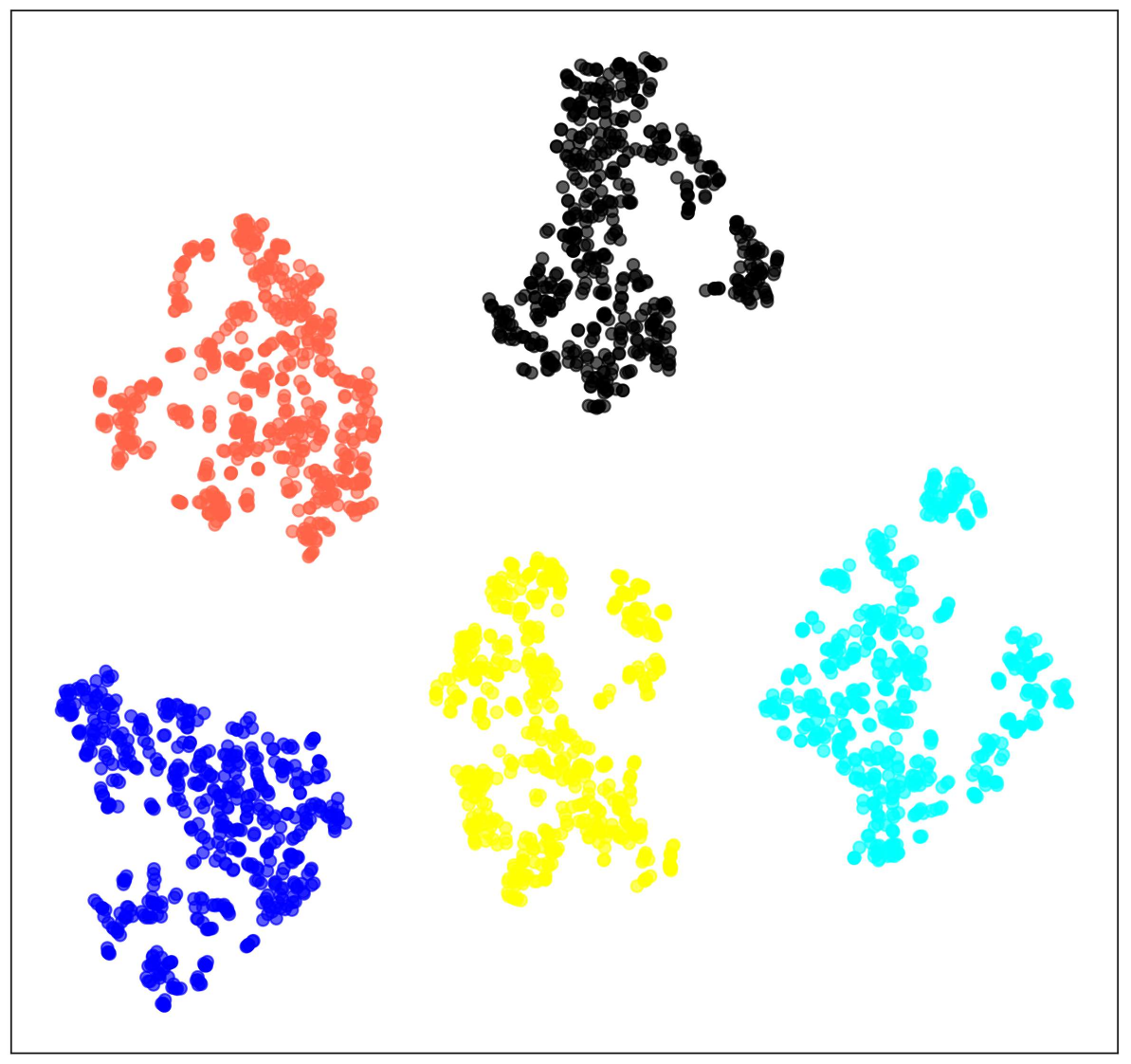}
    }
    \caption{The visualization of prototype feature distribution learned by (1) TRX vs.(2) STRM vs. (3) TSA-MLT on support samples for Kinetics under 5-way 5-shot settings. The visualization is performed with T-SNE.
    }
 \label{fig:7}
\end{figure}

\begin{figure}[htbp]
    \centering
    \subfloat[TRX prototype]{
    \label{fig:8.a}
    \includegraphics[width=0.16\textwidth]{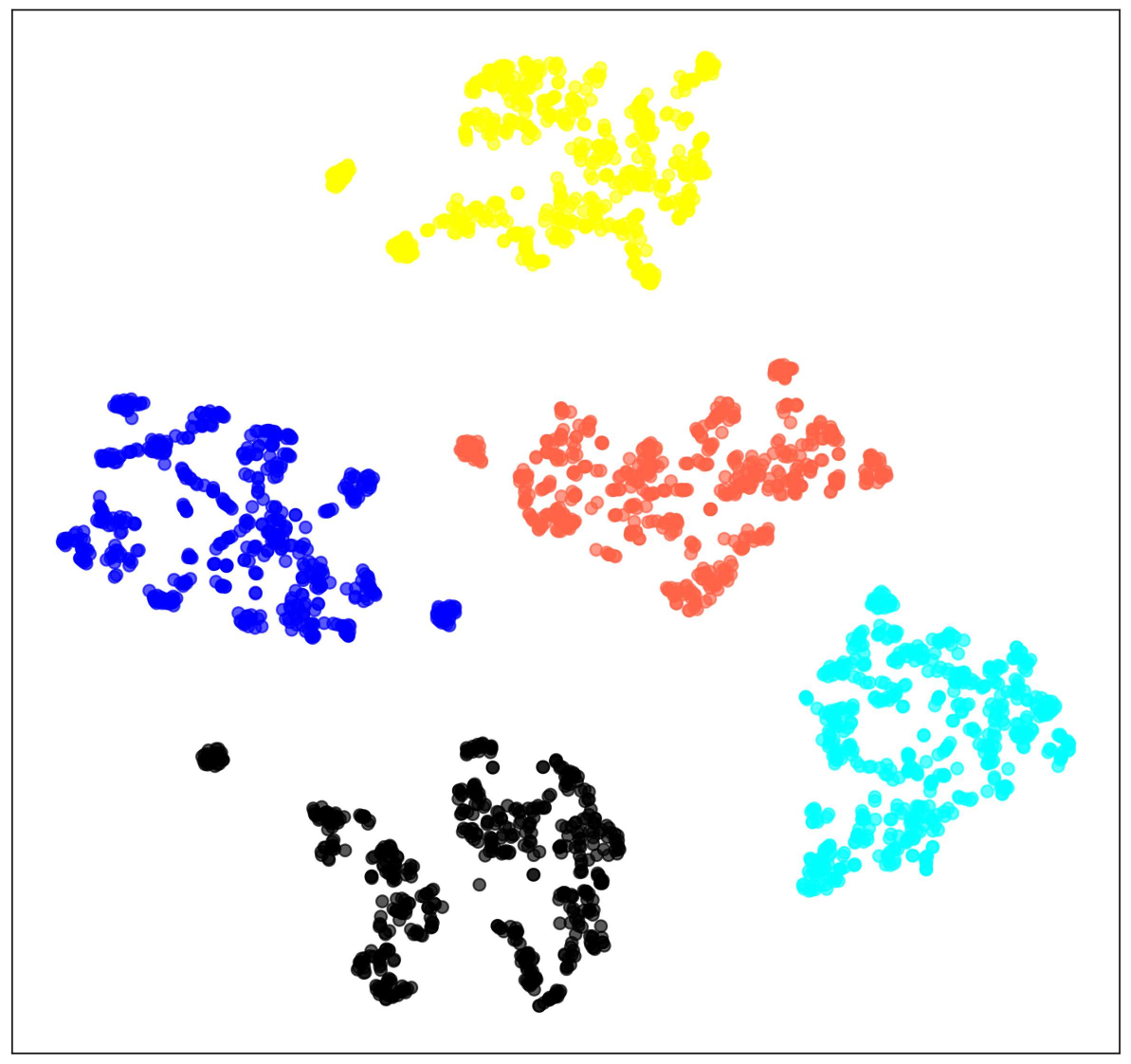}
    }
    \subfloat[STRM prototype]{
    \label{fig:8.b}
    \includegraphics[width=0.16\textwidth]{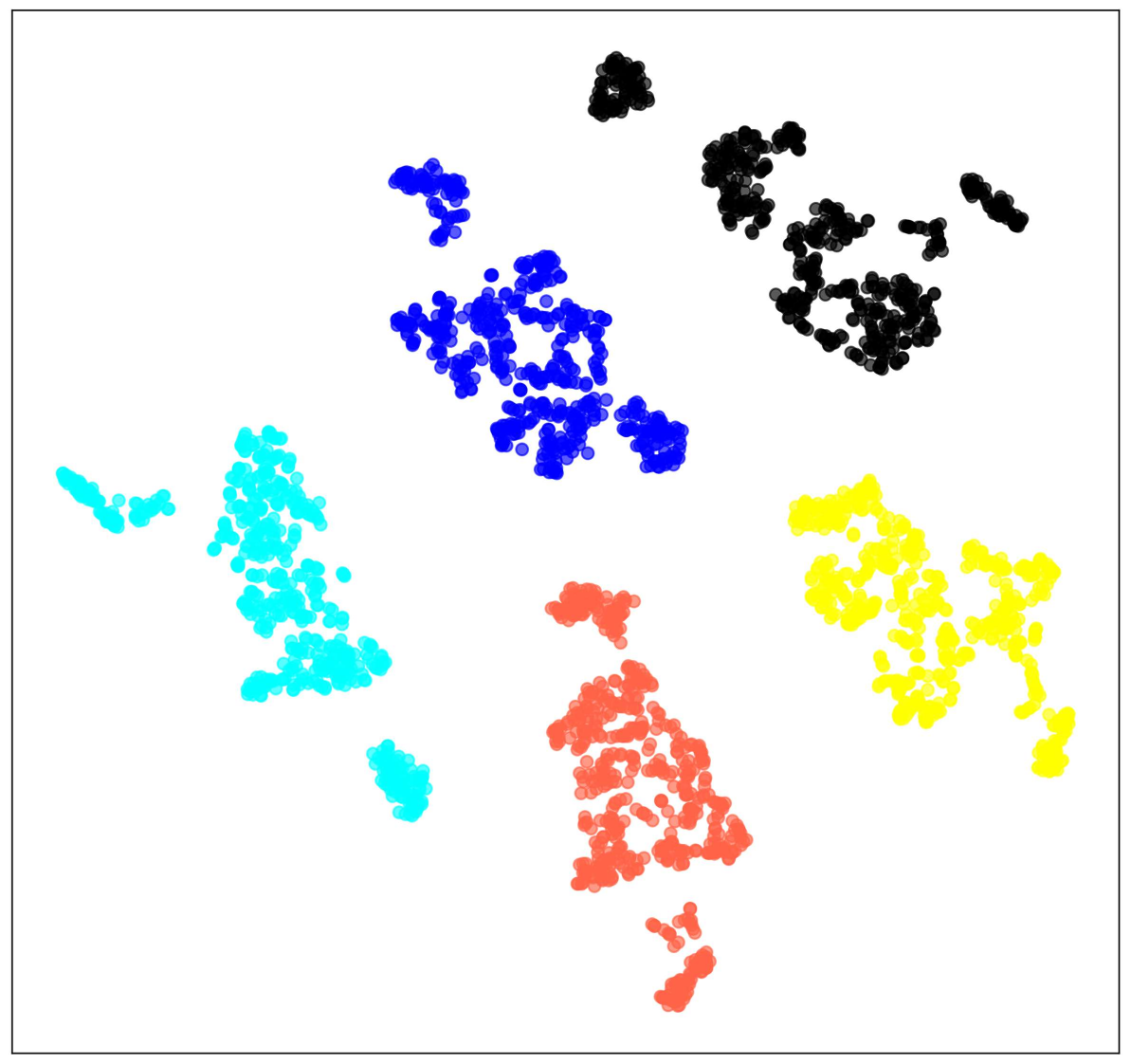}
    }
    \subfloat[TSA-MLT prototype]{
    \label{fig:8.c}
    \includegraphics[width=0.16\textwidth]{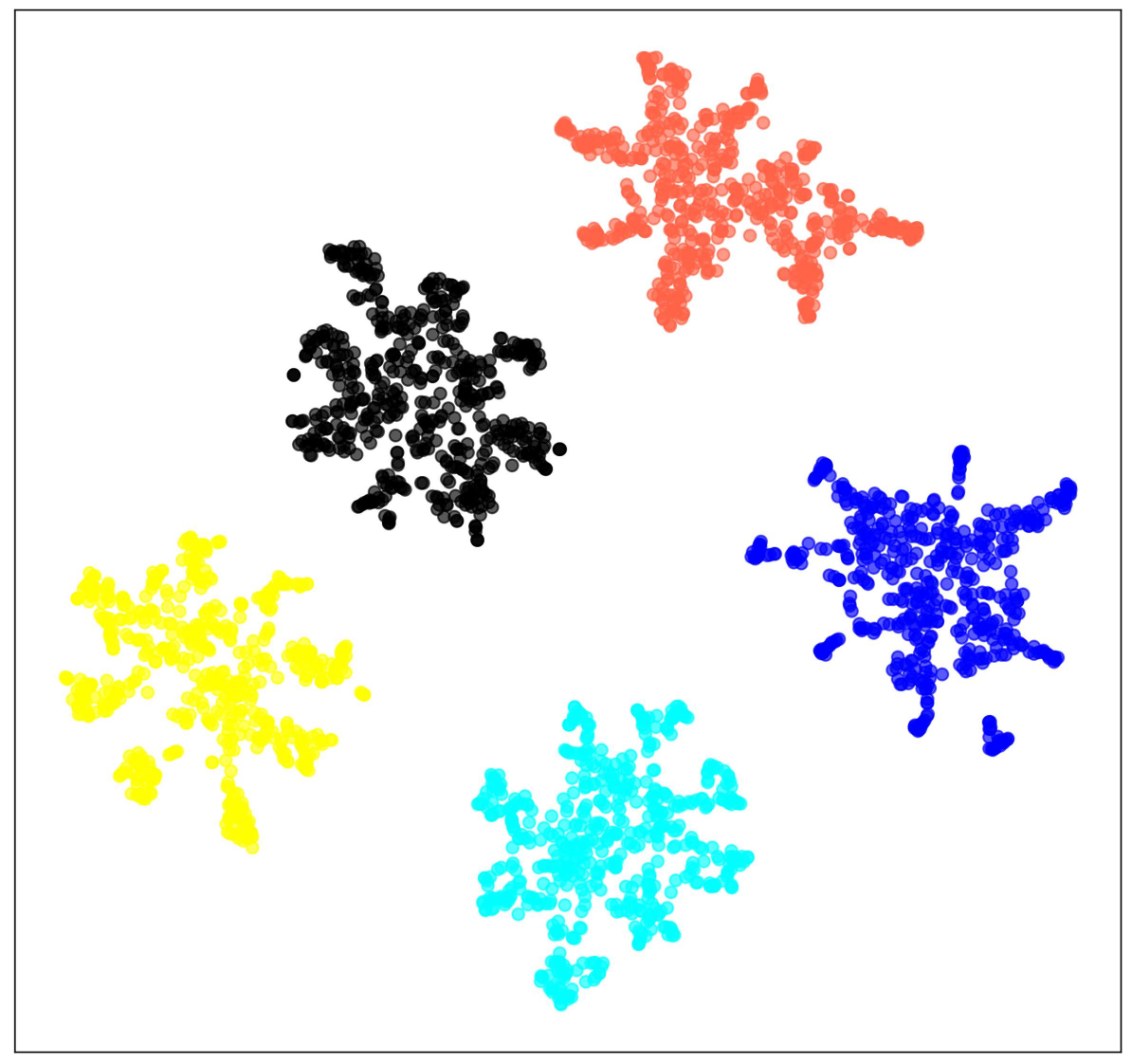}
    }
    \caption{The visualization of prototype feature distribution learned by (1) TRX vs.(2) STRM vs. (3) TSA-MLT on support samples for HMDB51 under 5-way 5-shot settings. The visualization is performed with T-SNE.
    }
 \label{fig:8}
\end{figure}

 To prove that our model can get better feature distribution, we visualize the feature distribution of the prototype learned by the TRX model, STRM model, and our TSA-MLT as shown in Figure \ref{fig:7} and \ref{fig:8} under 5-way 5-shot setting using t-SNE\cite{van2008visualizing-tsne} on Kinetics and HMDB51. The illustrations show that the feature in different classes is more discriminative, that the between-class scatter is larger, and the within-class scatter is smaller using TSA-MLT than other methods for both datasets. 

\subsection{Analysis of multiple feature attention score}

\begin{figure} 
\centering
\includegraphics[width=9cm]{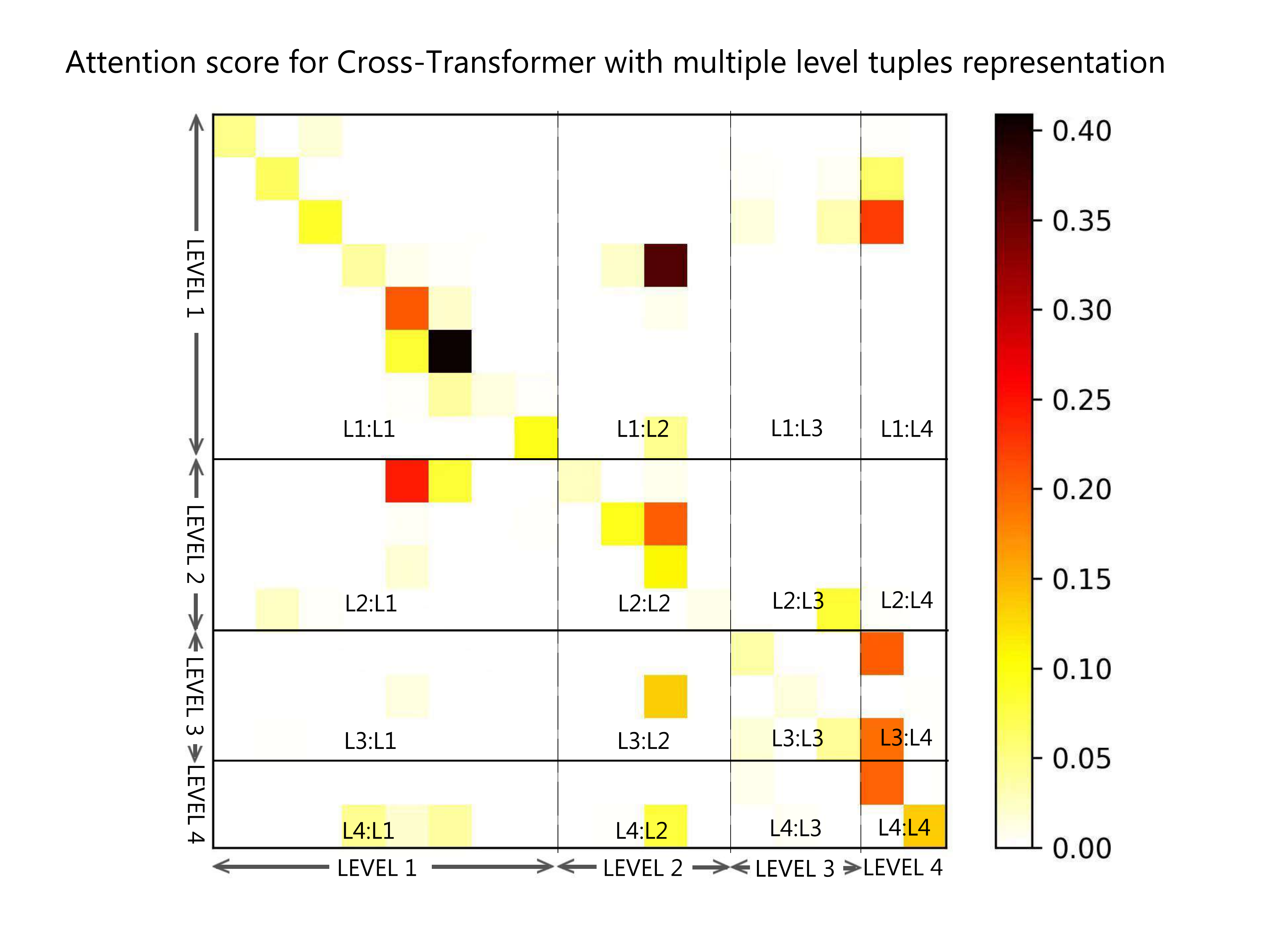}
\caption{\label{fig:attention_score}The illustration shows the attention score map for Cross-Transformer with Multiple-level tuples.}
\label{fig:9}
\end{figure}
In Figure \ref{fig:9}, we can see the attention score map for Cross-Transformer with Multiple-level tuples. Each row represents the query features, and each column represents the support features. The area of ``L1:L1" is the sub-attention score map for between level one representation, each row and column is the representation of a single frame in query features and support features, and the area of ``L1:L2" is the sub-attention score map between level one representation in query features and the level two representation in support features, each row is the representation of a single frame, and each column is the representation of the tuples that contain the information of 2 frames. Other areas also have a similar explanation. We can see there is a similarity between features not only at the same level but also at different levels. It makes sense that we use this multiple tuples attention map and the support features to compute the prototype in our paper, and each feature in the prototype contains information from different levels.

\subsection{Analysis of Optimal Transport matrix \texorpdfstring{$P^*$}{} }
\begin{figure} 
\centering
\includegraphics[width=9cm]{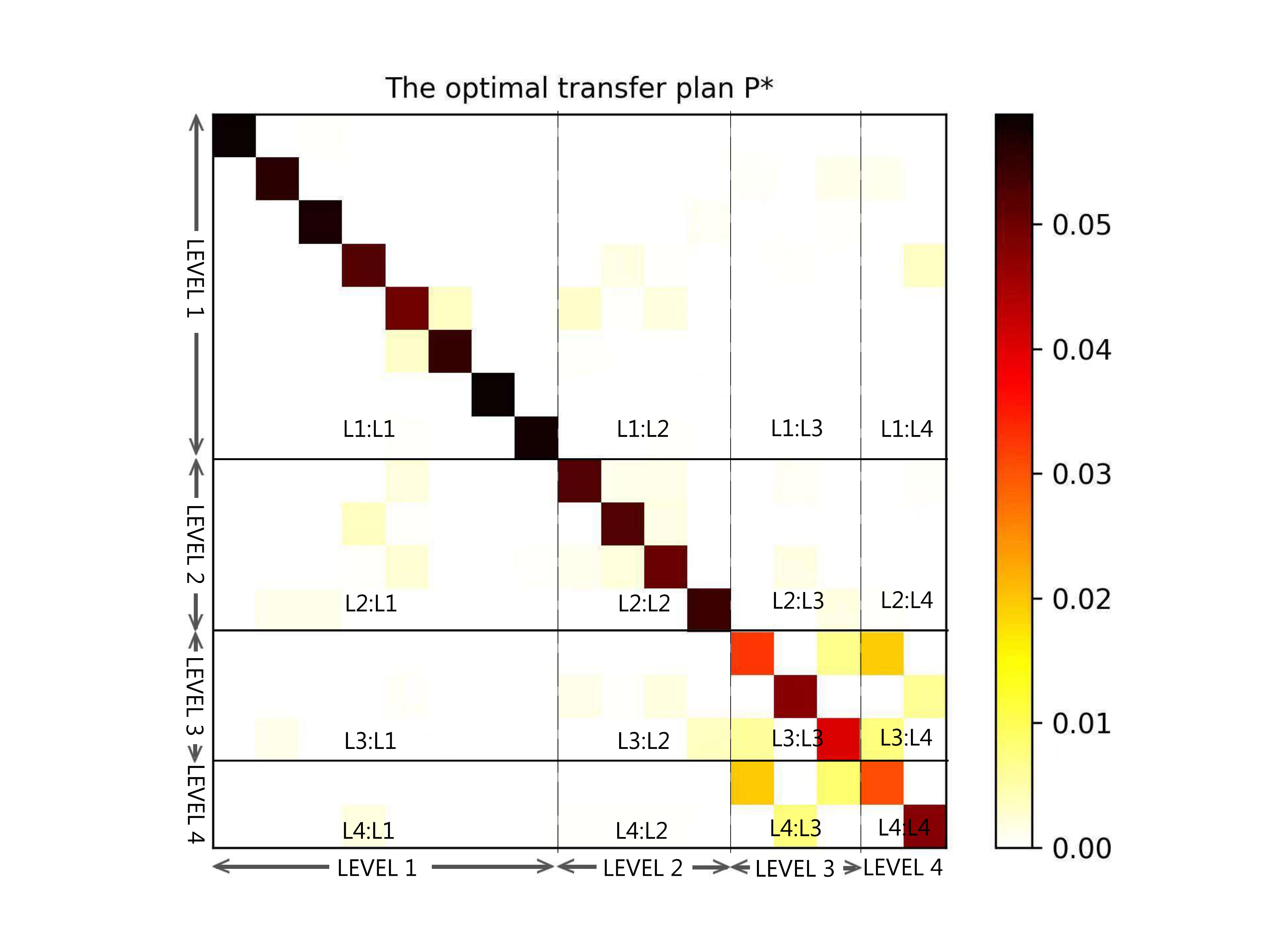}
\caption{\label{fig:transfer_plan}The illustration shows an Optimal Transport matrix $P^*$ of the Optimal Transport algorithm.}
\label{fig:10}
\end{figure}

In Figure \ref{fig:10}, it is the optimal transfer plan matrix $P^*$; for example, the value in row $i$ and column $j$ is the probability of transferring distribution between $ith$ query feature and the $jth$ prototype feature. It can be seen that a diagonal has a significant probability of transferring. The area of ``L1:L1" shows the transfer probability between the level one representation of the query and the level one representation of the query prototype, and the area of ``L1:L2" shows the transfer probability between the level one representation of the query and the level two representation of the query prototype. And all the other areas show the transfer probability between the same level or different levels. Figure \ref{fig:10} shows that features do have similarities not only from the same level but also from different levels.

\section{Limitation}
Through the experiments, it can be seen that our method has a certain competitiveness in accuracy compared to other methods. However, under the meta-learning rule, for each episode, we may not find the optimal cardinalities combination and the number of selected tuples created by the simple networks under each cardinality, which can explore its maximum potential. There are too many combinations, so how to scientifically select these combinations and tuples is worth exploring in the future. In addition, for one-shot learning, the accuracy of our method is somewhat lower than the current SOTA, which may be due to the limitations of the method for selecting tuples under cardinality. How to solve this problem is also worth exploring.

\section{Conclusion}
This paper proposes Task-Specific Alignment and Multiple-level Transformers for few-shot action recognition.
Firstly, we design a TSA module that uses a 3DCNN to get the parameter for affine warping and a 2DCNN for task-specific adjustment, which are used to filter some less important or semantically misleading frames. Secondly, we design a Multiple-level Transformer that gets several level semantic features of the video, from a single frame to the combinations of several frames. With this module, we can acquire different levels of semantic information about the video. Specifically, we use a linear network to decrease the number of tuples instead of using all the tuples under each level. Also, we design the Optimal Transport distance for Multiple-level features as a supplement to sequence distance. In the end, we use a fusion network to fuse the Sequence distance and Optimal Transport distance, then acquire a fusion distance for computing the final loss.

\section{Acknowledgements}
This work was supported by the Guangdong Basic and Applied Basic Research Foundation (Grant No. 2021A1515011913).

\bibliographystyle{cas-model2-names}
\bibliography{TSA-MLT}
\end{document}